\documentclass[journal,article,submit,pdftex,moreauthors]{Definitions/mdpi}

\firstpage{1} 
\pubvolume{1}
\issuenum{1}
\articlenumber{0}
\pubyear{2023}
\copyrightyear{2023}
\datereceived{ } 
\daterevised{ } 
\dateaccepted{ } 
\datepublished{ } 

\hreflink{https://doi.org/}
\pdfoutput=1

\Title{
Utilizing VQ-VAE for End-to-End Health Indicator Generation in Predicting Rolling Bearing RUL}

\TitleCitation{Utilizing VQ-VAE for End-to-End Health Indicator Generation in Predicting Rolling Bearing RUL}

\Author{Junliang Wang $^{1,2}$\orcidA{}, Qinghua Zhang $^{3,}$*\orcidB{} , Guanhua Zhu$^{3,}$*\orcidC{}  and Guoxi Sun$^{3}$}

\AuthorNames{Junliang Wang,  Qinghua Zhang, Guanhua Zhu and Guoxi Sun}

\AuthorCitation{Wang, J.; Zhang, Q.; Zhu, G.; Sun, G.}

\address{%
$^{1}$ \quad School of Automation, Guangdong University of Petrochemical Technology, Maoming 525000, China;
2112204457@mail2.gdut.edu.cn\\
$^{2}$ \quad  School of Automation, Guangdong University of Technology, Guangzhou 510006, China; \\
$^{3}$ \quad Guangdong Provincial Key Laboratory of Petrochemical Equipment and Fault Diagnosis,
Guangdong University of Petrochemical Technology, Maoming 525000, China;  guoxi.sun@gdupt.edu.cn
}

\corres{Correspondence: fengliangren@tom.com; zhugh@gdupt.edu.cn; }

\abstract{The prediction of the remaining useful life (RUL) of rolling bearings is a pivotal issue in industrial production. A crucial approach to tackling this issue involves transforming vibration signals into health indicators (HI) to aid model training. This paper presents an end-to-end HI construction method, vector quantised variational autoencoder (VQ-VAE), which addresses the need for dimensionality reduction of latent variables in traditional unsupervised learning methods such as autoencoder. Moreover, concerning the inadequacy of traditional statistical metrics in reflecting curve fluctuations accurately, two novel statistical metrics, mean absolute distance (MAD) and mean variance (MV), are introduced. These metrics accurately depict the fluctuation patterns in the curves, thereby indicating the model's accuracy in discerning similar features. On the PMH2012 dataset, methods employing VQ-VAE for label construction achieved lower values for MAD and MV. Furthermore, the ASTCN prediction model trained with VQ-VAE labels demonstrated commendable performance, attaining the lowest values for MAD and MV.}

\keyword{RUL;  rolling bearing;  health indicator; Autoencoder; VQ-VAE } 

\begin{document}





\section{Introduction}
Rolling bearings, often referred to as "the joints of the industry," are crucial components of rotating machinery and find extensive applications across various industrial domains\cite{ref1,ref2,ref3}. The least severe outcome of a bearing failure is the malfunction of equipment and subsequent economic losses; the most severe is the initiation of safety incidents and infliction of personal injuries. Consequently, the ability to predict a bearing's remaining useful life (RUL) with precision and promptness is of considerable research interest in the domain of industrial safety.
Generally speaking, RUL predictions can be categorized into model-based, data-driven, and hybrid methods\cite{ref4}. Approaches based on models employ physical representations of the degradation trajectory, integrating these with empirical data to project the forthcoming decline. For example, Singleton et al.\cite{ref5} introduces a method utilizing an extended Kalman filter predicated on continuous-time state equations to estimate RUL. Qian et al.\cite{ref6} presents an enhanced particle filter technique that uses a variable importance density function and a backpropagation neural network to increase particle diversity for accurate prediction of RUL.

The acquisition of physical models can become a formidable challenge in the face of complex systems or elaborate operating conditions. In such contexts, the abundance of monitoring data paves the way for the application of data-driven methods\cite{ref7,ref8,ref16,ref17,ref18,ref19,ref20,ref21} that harness artificial intelligence algorithms for the prediction of RUL. These methods are characterized by their lack of reliance on prior knowledge, enabling their flexible deployment across a spectrum of problems and situational contexts. As a result, when a substantial dataset related to operational failures is at hand, predictive efforts tend to pivot primarily towards data-driven methodologies. 

Traditional machine learning methods include neural networks, support vector machines, relevance vector machines, bayesian inference, and fuzzy systems\cite{ref9,ref10,ref11,ref12,ref13}. These types of mapping models are usually achieved by conducting regression analysis to model the relationship between RUL and monitored data. Besides the aforementioned methods, hybrid approaches aim to integrate physical models with data-driven methods, amalgamating the advantages of both while mitigating their limitations concerning RUL prediction\cite{ref14}. For instance, Cheng et al.\cite{ref15} combines the adaptive neuro-fuzzy inference system with particle filters, utilized respectively for learning the state transition functions and predicting RUL based on the learned state transition functions.

Over several years of development, deep learning methods have achieved breakthroughs in many fields including image classification, speech recognition, and natural language understanding. Even in the realm of fault diagnosis and prediction, deep learning-based methods for bearing RUL prediction have gradually become mainstream. Through the exploitation of large datasets and advanced computational capabilities, deep learning facilitates the extraction of complex patterns and features from data, thereby significantly enhancing the accuracy and reliability of RUL predictions in comparison to traditional machine learning approaches. Ren et al.\cite{ref16} proposes a new method for the prediction of bearing RUL based on deep convolution neural network (CNN). Guo et al.\cite{ref22} study a recurrent neural network (RNN) based health indicator for RUL  prediction of bearings. As an improvement of the RNN, long short-term memory (LSTM) based neural network was introduced by Yuan et al.\cite{ref17} for effective fault diagnosis and RUL estimation of aero engines. Que et al.\cite{ref18} study remaining useful life prediction for bearings based on a gated recurrent unit (GRU), and an ensemble data-driven approach is proposed to predict the RUL of bearings.

The temporal convolutional network (TCN) is a recent improvement on the CNN architecture, utilizing dilated causal convolution (DCC) to extract information from historical data. Dilated causal convolution typically comprises fewer layers compared to classical CNNs, yet it is capable of capturing the same receptive field. Through the dilation operation, TCN expands the kernel's reach without requiring additional parameters, hence maintaining computational efficiency while still being able to discern long-term dependencies in the data, making it a suitable choice for time-sensitive applications such as RUL prediction and time series forecasting\cite{ref19,ref20,ref21}. Therefore, this paper adopts TCN as the backbone network for the experiments.

Additionally, it is common practice now to first convert vibration signals into health indicators (HI), which can guide the network in various ways. Kong et al.\cite{ref23} uses the mean absolute value of signal extremes (MAVE) to characterize the signal energy to solve the problem that the HI selected by the existing model at the first prediction time (FPT) is insensitive to the initial failure.  Nowadays, the method of utilizing latent values through the hidden layer of an autoencoder (AE) to create HI is becoming increasingly popular. Chen et al.\cite{ref24} introduces a novel deep convolutional autoencoder based on a quadratic function for constructing health indicators from raw vibration signals to predict the RUL of bearings. Liu et al.\cite{ref25} study densely connected fully convolutional auto-encoder based slewing bearing degradation trend prediction method. The proposed method is verified by large-scale slewing bearing data from the highly accelerated life test. Despite the increasingly mature techniques in creating HI, there are two shortcomings: (1) some methods perform well on specific datasets but lack generalization capabilities, leading to decreased performance when applied to different types of bearings or under varying operating conditions; (2) extracting useful features from raw vibration signals is challenging, and existing methods may not adequately capture all the information related to bearing health contained within the signals.

\begin{figure}
    \centering
    \includegraphics[width=1.0\linewidth]{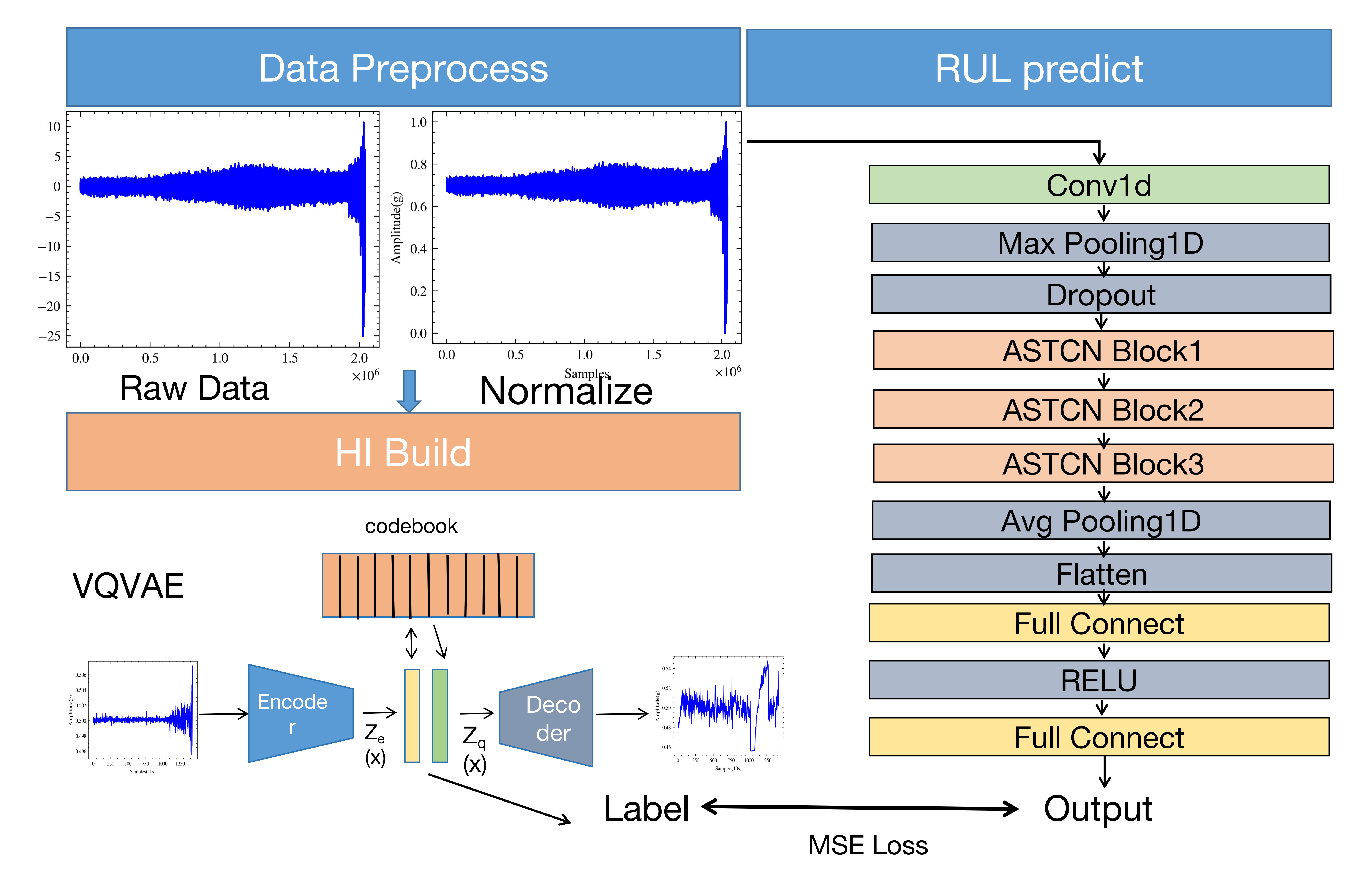}
    \caption{Proposed RUL prediction method.}
    \label{fig:flow}
\end{figure}

In order to solve the issues in health indicator generation, this paper establishes a new RUL prediction framework, with the specifics of this method illustrated in Figure \ref{fig:flow}.
Upon normalization of all data, it is fed into VQ-VAE for label generation. As an improvement over the AE, the VQ-VAE\cite{ref29} was initially developed for image generation tasks, providing a means to learn discrete latent representations, which are useful for generating high-quality images. The introduction of a discrete codebook in VQ-VAE enables the model to handle more complex data distributions than standard AEs, making it particularly powerful for tasks that benefit from compressed, yet expressive, representations of data. The created labels along with the normalized data are then sent to the predictive neural network TCN for training. The loss of the model is computed as the mean squared error (MSE) between the predicted results and the labels obtained through VQ-VAE.
The contributions of this paper can be summarized in three main points:
\begin{enumerate}
\item An end-to-end HI construction method, VQ-VAE, is introduced for the prediction of bearing RUL. This method optimizes the shortcomings of HI metrics generated by traditional methods and resolves the need for dimensionality reduction of latent variables in conventional unsupervised learning methods such as Autoencoder.
\item In light of the inadequacy of traditional statistical metrics like MSE, MAE, Score, etc., in reflecting the fluctuation in curves accurately, two novel statistical metrics, mean absolute distance (MAD) and mean variance (MV), are proposed. These metrics can accurately depict the fluctuation patterns in curves, thereby reflecting the model's accuracy in identifying similar features.
\item  A novel RUL prediction framework is established, constituting an end-to-end prediction process that directly employs vibration signal inputs, eliminating the need for a priori knowledge for feature selection and dimensionality reduction.
\end{enumerate}

This paper is structured as follows: Section \ref{section:pre} briefly describes the basic principles of temporal convolutional networks (TCN), autoencoder (AE), and vector quantised variational autoencoder (VQ-VAE). Section \ref{section:model} introduces the basic network structures of ASTCN, AE, and VQ-VAE, as well as the associated label generation methods. Section \ref{section:exp} evaluates the practical performance of the RUL prediction framework through experiments. Section \ref{section:con} concludes the paper and proposes some directions for future work.
\section{Preliminary}\label{section:pre}
\subsection{Dilated Causal Convolution (DCC)} 
Causal convolution is a type of convolutional operation used primarily in the context of time series data and sequence modeling, where the output at a given time is determined only by the current and past inputs, not by any future inputs. To achieve this, causal convolutions often use padding on the left side (past) of the input sequence but not on the right side (future), ensuring that the convolutional filter does not see future data\cite{ref26}. Dilated causal convolution is an extension of causal convolution.
The dilation aspect introduces a skip factor in the convolution operation, allowing the network to encompass a broader scope of input data without the necessity for increasing the number of parameters or layers, thus maintaining computational efficacy.
Formally, the dilated convolution operation can be defined as follows:
\begin{figure}
    \centering
    \includegraphics[width=1\linewidth]{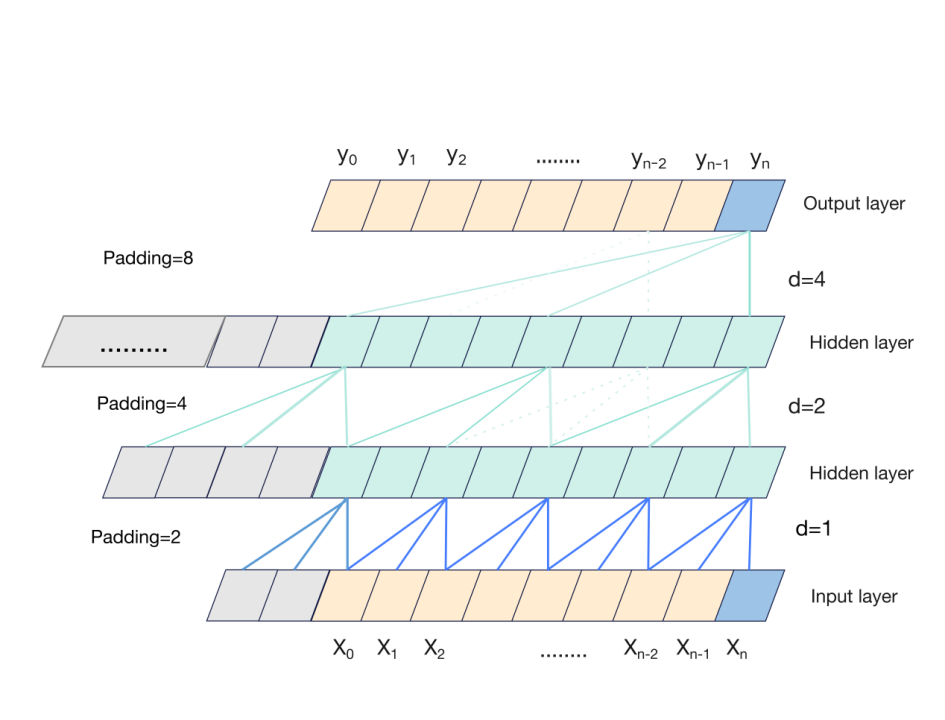}
    \caption{The DCC architechure of TCN. \label{fig1}} 
\end{figure}

\begin{equation}
y(t)=\sum_{i=0}^{k-1}f(i)\cdot x(t-d\cdot i)
\end{equation}
where $y(t)$ is the output at time t, $x(t)$ is the input at time t, $f(i)$ is the filter value at position i, $k$ is the size of the filter, and $d$ is the dilation factor.

In traditional convolution, the dilation factor d is 1, meaning that the filter operates on consecutive input values. However, with a dilation factor greater than 1, the operation skips $d-1$ positions between inputs, thereby extending the receptive field of the convolution without enlarging the filter size, as the Figure ~\ref{fig1} shows.
The causal aspect ensures that the output at any time $t$ is only dependent on the current and past input values, adhering to the temporal ordering and guaranteeing that future data does not influence the present output. This is crucial for tasks such as time series forecasting or RUL prediction\cite{ref19,ref20,ref21}, where the model's predictions should only be influenced by past and present data.

\subsection{Residual} 
In TCNs, the depth of the network directly affects the size of its receptive field, thereby influencing the model's ability to learn and understand long-term dependencies in time series data. The utilization of residual connections\cite{ref27} allows for the training of significantly deeper networks without encountering the challenges of increased training difficulty or the vanishing gradient problem. Within the TCN architecture, the incorporation of residual connections fosters a robust and efficient learning environment, especially over extended temporal sequences, ensuring that crucial information from earlier time steps is carried forward and not lost, thus substantively aiding in capturing long-term dependencies inherent in sequential data.

\begin{figure}
    \includegraphics[width=0.5\linewidth]{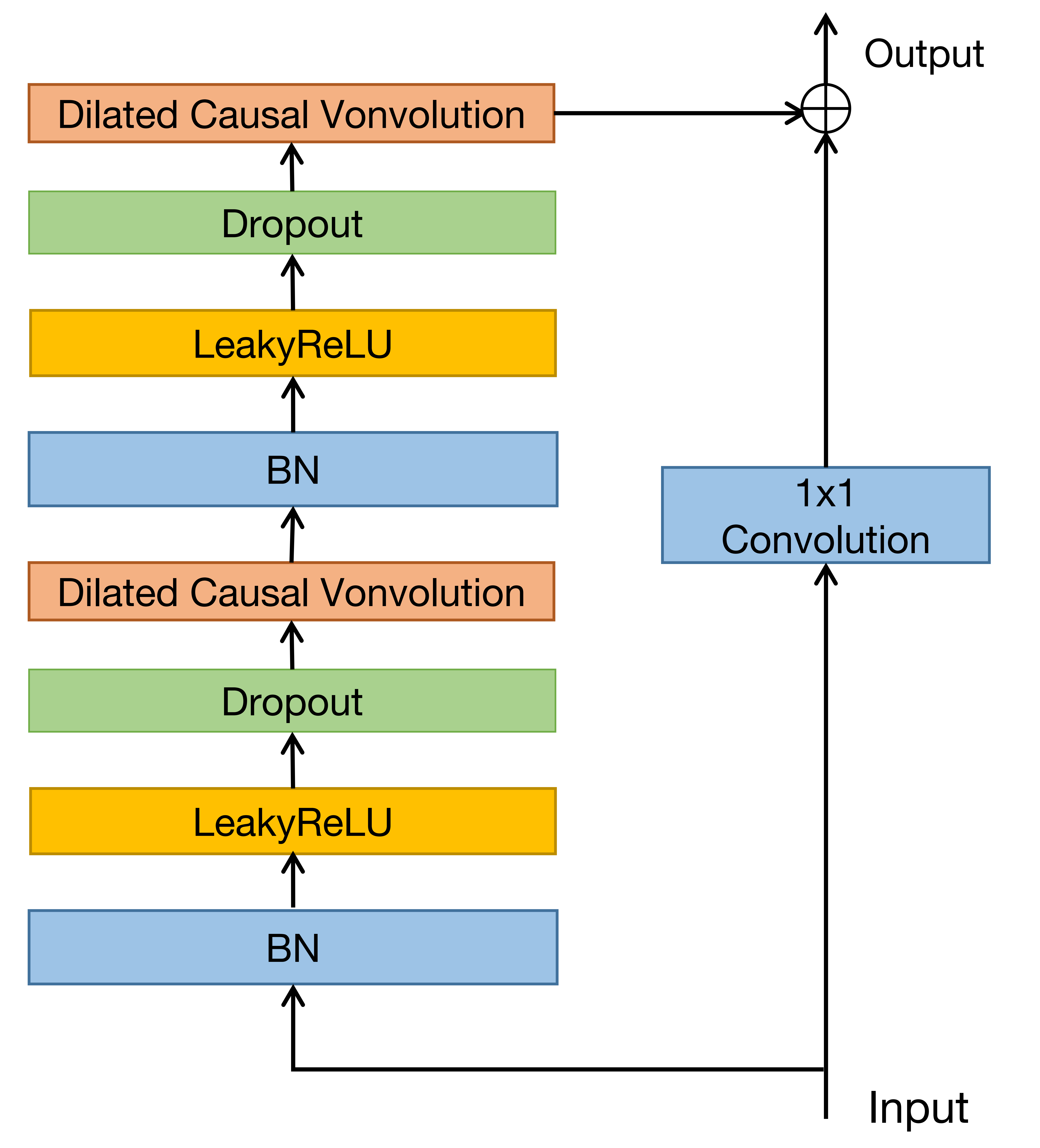}
    \caption{The block diagram of the fully residual module. \label{fig2}}
    \label{fig:enter-label}
\end{figure}

These connections, also known as skip connections, form shortcuts that bypass one or more layers in the network, thus facilitating an easier learning process and mitigating the common vanishing gradient problem encountered in deep networks. The residual connection is represented as
$Y=F(X)+X$, where $Y$ is the output, $X$ is the input, and $F(X)$ denotes the transformation applied by the layer, including operations such as convolution and activation functions. The improved residual structure is presented in Figure ~\ref{fig2}. Each TCN block consisting of two dilated causal convolution blocks, preceded by batch normalization, LeakyReLU  and dropout  and the final output is the sum of the dilated causal convolution and the 1x1 convolution.

\subsection{VQ-VAE}
An autoencoder (AE) is a variant of artificial neural networks designed for the unsupervised learning of efficient encodings.The primary goal of an AE is to learn a representation (encoding) for a set of data, typically for dimensionality reduction\cite{ref28}. An autoencoder is structured into two key segments: the encoder, which condenses the input into a condensed latent representation, and the decoder, which strives to recreate the initial input from this compressed form. The fundamental principle of an AE's operation is to diminish the variance between the original data and its reconstructed counterpart. This reduction in the reconstruction discrepancy is facilitated by employing backpropagation and gradient descent techniques to refine the network's parameters and minimize the error in the output. Mathematically, this can be represented as minimizing the loss function $L(x,\hat x)$, where $x$ is the input data and $\hat x$ is the reconstructed output.
Formally, the encoding and decoding processes in an autoencoder are expressed as:

\begin{equation}
\begin{aligned}
h&=f(W\cdot x+b)\\
\hat{x}&=g(W'\cdot h+b')
\end{aligned}
\end{equation}
where  $h$ is the encoded representation, $f$ and $g$ are the activation functions for the encoding and decoding processes respectively, $W$ and $W'$ are the weight matrices, $b$ and $b'$ are the bias vectors.

\begin{figure}
    \centering
    \includegraphics[width=1\linewidth]{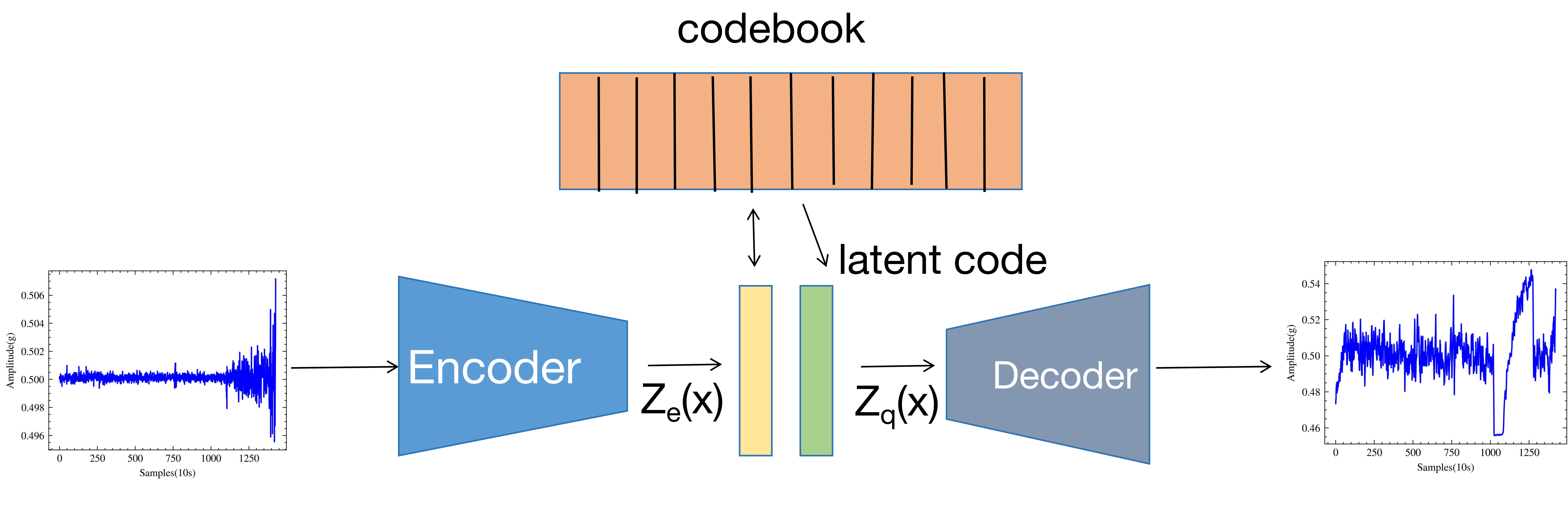}
    \caption{The schematic diagram of VQ-VAE.}
    \label{fig:vqvae-block}
\end{figure}

Building upon the fundamental principles of AE, VQ-VAE\cite{ref29} introduces a discrete latent representation, aiming to address the limitations of continuous latent variables often encountered in variational autoencoders (VAEs) \cite{ref30}. The core idea behind VQ-VAE is to employ vector quantization in the latent space, wherein each point in the latent space is mapped to the nearest point in a predefined grid of embeddings, as Figure \ref{fig:vqvae-block} shows.

The quantization process in VQ-VAE is achieved by employing a codebook of vectors, and each element in the latent space is replaced by the nearest code in the codebook. The objective function of VQ-VAE comprises three terms: a reconstruction loss term ensuring the quality of reconstruction,  a codebook loss ensuring the effective utilization of the codebook, and a commitment loss term encouraging the encoder to commit to a particular code, as the Equation (3) shows:

\begin{equation}L=\log p(x|z_q(x))+\|sg[z_e(x)]-e\|_2^2+\beta
\|z_e(x)-sg[e]\|_2^2
\end{equation}
where  $x$ is the input value, and 
 $z_q(x)$ is its reconstructed counterpart via the decoder. The term 
$e$ represents the quantized output from the codebook, while  $z_e(x)$
is the encoder's output. The 'sg' operation denotes the 'stop gradient' process. Lastly, $\beta$ serves as a coefficient that weights the commitment loss in the model.

$logp(x\mid z_q(x))$ represents the reconstruction loss. It quantifies the divergence between the actual input $x$ and its reconstructed version based on the quantized representation $z_q(x)$. The purpose of this term is to ensure a faithful reproduction of the input post-encoding and decoding.

$\left\lVert sg[z_e(x)] - e \right\rVert_2^2$  corresponds to the codebook loss. The 'sg' denotes the 'stop gradient' operation, which prevents certain gradients from updating during the backpropagation process. This loss component ensures that the encoder's continuous output closely matches the discrete codebook entries.

$\beta\left\lVert z_e(x) - sg[e]\right\rVert_2^2$  is the commitment loss. The coefficient $\beta$ denotes the weight of this term, emphasizing the importance of the encoder's commitment to specific codebook vectors. The 'sg' applied to $e$ ensures that during optimization, the encoder's weights are updated based on this loss, while the codebook remains unchanged.

\section{Models and Labels}\label{section:model}
\subsection{Model structures}
\subsubsection{TCN}

Compared to some feature extraction methods, using raw data directly can better reflect the network's feature extraction capability\cite{ref40}. Therefore, we chose ASTCN
\cite{ref31} as our backbone network.

\begin{figure}
    \centering
    \includegraphics[width=1.0\linewidth]{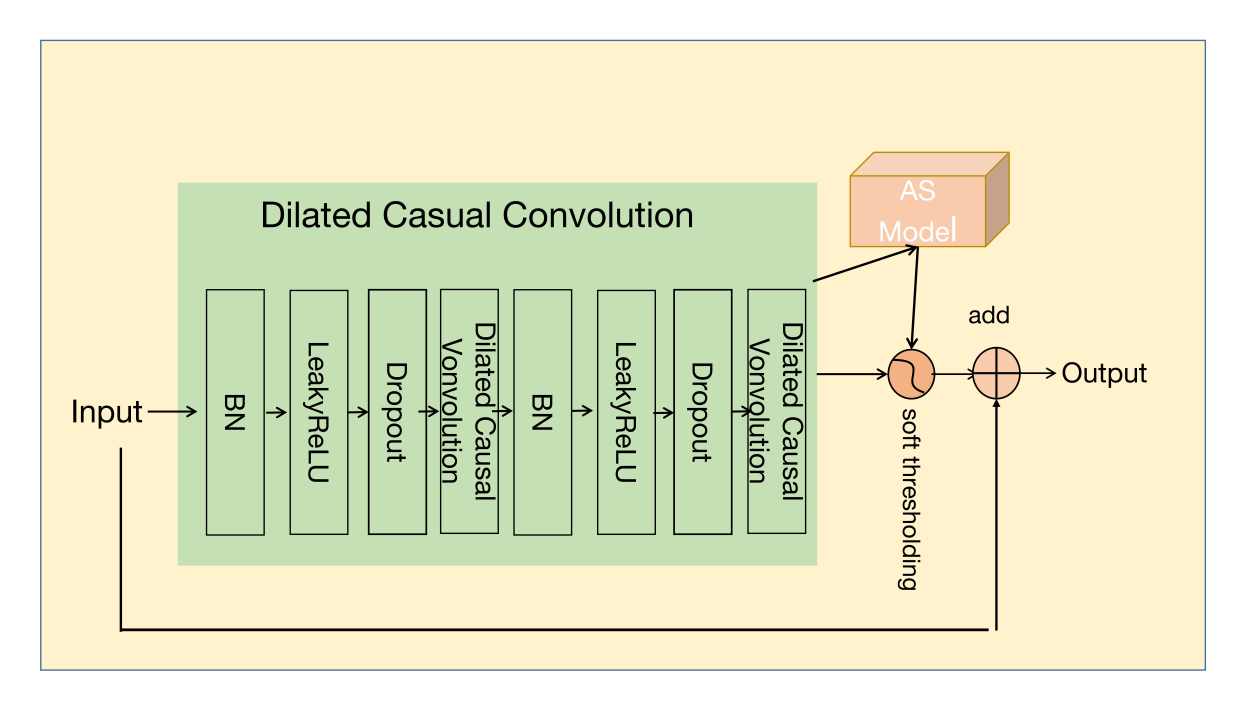}
    \caption{The ASTCN block structure.}
    \label{fig:astcn-block}
\end{figure}

As the Figure \ref{fig:astcn-block} shows, the ASTCN's block, apart from the TCN block, also includes the AS module and the soft thresholding function. The AS module takes in a feature map and calculates the threshold for each channel, which is then inputted into the soft thresholding function. Its working mechanism is as follows: Suppose that the input tensor has $N$ rows and $M$ columns; the tensor undergoes an absolute operation, followed by global average pooling. Subsequently, we obtain a tensor $h$ with dimensions $N \times 1$. This tensor is then successively passed through a fully connected layer, a BN layer, and a sigmoid activation function to constrain its values between 0 and 1. This result is named as $h'$. By performing element-wise multiplication between $h$ and $h'$, we obtain the final result with $N$ rows and $1$ column, meaning that each channel gets a corresponding threshold value. The boundary in the soft thresholding function is controlled by a threshold $\tau$ which is actually computed by the AS Model. Then, by adding the value initially inputted into the block, we obtain the block's output.

The soft thresholding function, given by Equation (4), is a fundamental tool utilized predominantly in the realm of sparse signal processing and compressive sensing. Given a threshold parameter $\tau$, this function serves to shrink the magnitude of its input $x$ towards zero. Specifically, for magnitudes less than or equal to $\tau$,  the function maps the input directly to zero, effectively introducing sparsity. On the other hand, for values exceeding $\tau$ in either positive or negative direction, the function decreases the absolute magnitude by $\tau$, ensuring that only significant coefficients persist while lesser important ones are suppressed. By using the soft thresholding function as an activation function, we can encourage the network to produce sparse activations. This means that in certain scenarios, the output of some neurons will be set to zero, which can help reduce the complexity of the network and enhance interpretability.

\begin{equation}
soft(x,\tau)=y=\begin{cases}x+\tau,x<-\tau\\0,|x|\leq\tau\\x-\tau,x>\tau\end{cases}
\end{equation}
where $y$ and $x$ are the output and input feature, respectively.

\begin{figure}
    \centering
    \includegraphics[width=0.75\linewidth]{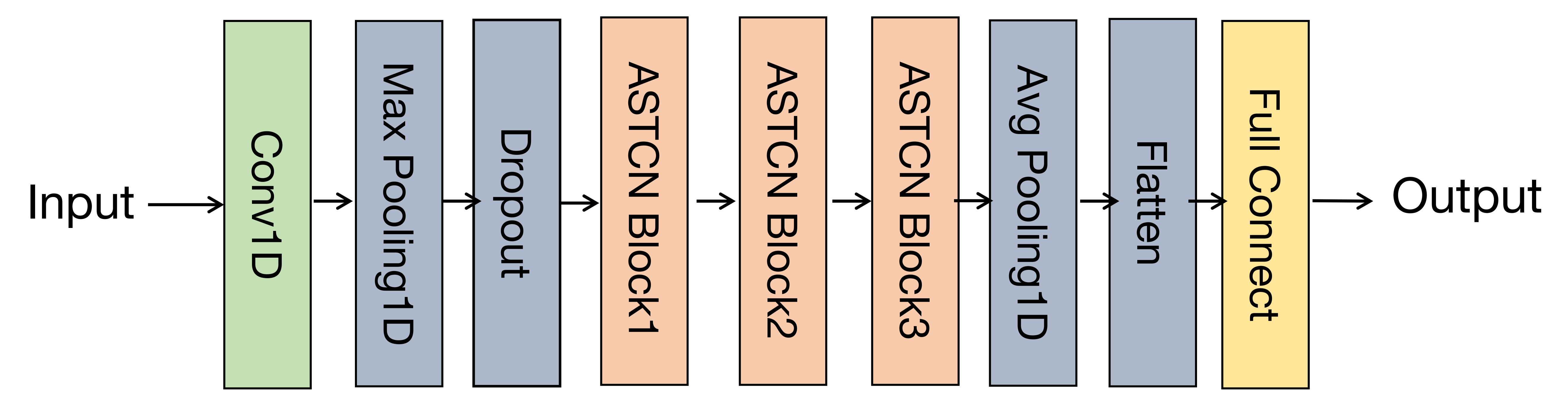}
    \caption{The ASTCN network structure. }
    \label{fig:astcn-model}
\end{figure}

The ASTCN\cite{ref31} structure is shown in Figure ~\ref{fig:astcn-model}, and structural parameters is shown in Table \ref{table:astcn}. The network starts with an input layer for time series data, followed by a convolutional layer and a max pooling layer. After the initial processing, the data passes through a dropout layer for regularization and then sequentially through three ASTCN blocks. Post-processing includes a global average pooling layer, a flatten layer, a full connection layer, a ReLU activation function, and a layer that outputs the predicted  RUL. The network calculates the loss by comparing this prediction to the labeled RUL and uses backpropagation for optimization during training.

\begin{table}[h]
\caption{The parameters of the ASTCN. \label{table:astcn}}
\begin{tabularx}{\textwidth}{cc}
\toprule
\textbf{Layer} & \textbf{Parameter} \\
\midrule
Conv 1D & Kernel length = 12, channel = 16, stride = 4 \\

Maxpooling 1D & Kernel length = 4, channel = 16, pooling size = 4 \\

Dropout & Dropout rate = 0.3 \\

AS-TCN Block 1 & Kernel length = 3, channel = 12, dilation rate = 1,  leaky rate = 0.2, dropout =0.3 \\

AS-TCN Block 2 & Kernel length = 3, channel = 6, dilation rate = 2,   leaky rate = 0.2, dropout = 0.3 \\

AS-TCN Block 3 & Kernel length = 3, channel = 4, dilation rate = 4,  leaky rate = 0.2, dropout = 0.3 \\

Dense & channel = 1 \\
\bottomrule
\end{tabularx}

\end{table}

\subsubsection{VQVAE}
We use AE \cite{ref32} to produce health indicators. Its structure is quite simple, as shown in Table \ref{tab:autoencoder}, The encoder consists of four convolutional blocks and another two 1x1 convolutions. Each block is composed of a convolution layer, BN layer, ReLU layer, and max pooling layer. 
The structure of the decoder is symmetrical to the encoder, placing the latter convolution at the forefront and replacing the max pooling layer in each block with an upsampling layer. The size of all convolution kernels in the autoencoder is 3x1.

\begin{table} \caption{The parameters of the VQ-VAE }  \begin{tabularx}{\textwidth}{XXXXX} \toprule Layer name & Kernels size & Stride & Kernels number & Output size \\ \midrule Input & – & – & – & 1024 x 1 \\ Convolution1 & 1 x 3 & 1 x 1 & 4 & 1024 x 4 \\ Pooling1 & 1 x 2 & 1 x 2 & – & 512 x 4 \\ Convolution2 & 1 x 3 & 1 x 1 & 4 & 512 x 4 \\ Pooling2 & 1 x 2 & 1 x 2 & – & 256 x 4 \\ Convolution3 & 1 x 3 & 1 x 1 & 8 & 256 x 8 \\ Pooling3 & 1 x 2 & 1 x 2 & – & 128 x 8 \\ Convolution4 & 1 x 3 & 1 x 1 & 8 & 128 x 8 \\ Pooling4 & 1 x 2 & 1 x 2 & – & 64 x 8 \\ Convolution5 & 1 x 3 & 1 x 1 & 16 & 64 x 16 \\  Pooling6 & 1 x 2 & 1 x 2 & – & 32 x 16\\ Convolution6 & 1 x 3 & 1 x 1 & 16 & 32 x 16  \\ \bottomrule \end{tabularx}  \label{tab:VQ-VAE} \end{table}
VQ-VAE\cite{ref29} and AE have fundamentally similar structures, but there is a change in the number of channels in the convolutional kernels, as shown in Table \ref{tab:VQ-VAE}. Moreover, VQ-VAE\cite{ref29} introduces an additional component called the codebook. The dimensions of the codebook are 32x16, meaning there are 32 embeddings, and each embedding is 16-dimensional.

In addition to directly receiving vibration data, the VQ-VAE and AE in this paper also accept 38-dimensional data that has undergone feature selection.
For handling 38-dimensional data, modifications in the network architecture have been made by reducing two convolutional layers along with their corresponding max-pooling layers, as shown in Table \ref{tab:small-autoencoder} and Table \ref{tab:small-VQVAE}. This reduction is aimed at maintaining a balance between the model's complexity and its capability to capture essential information from the data.

\subsection{Label Introduction}
\begin{figure}
    \centering
    \includegraphics[width=1.0\linewidth]{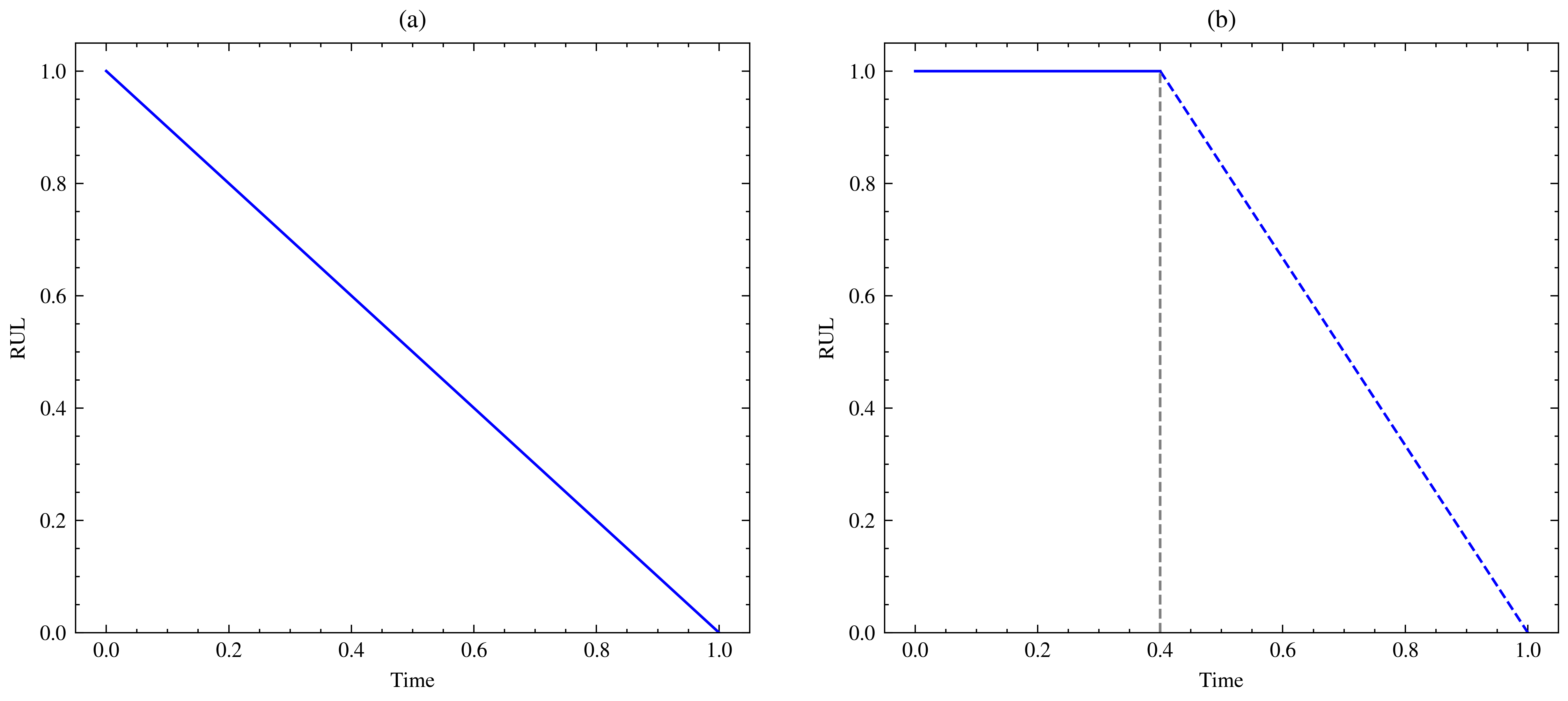}
    \caption{Two basic function curves: (a)linear function. (b) piecewise function.}
    \label{fig:curve1.png}
\end{figure}

Currently, most remaining life prediction problems still adopt supervised learning training methods. This means that the ability to accurately label data directly affects the final prediction accuracy. There are two primary methods to add labels \cite{ref33,ref34,ref35,ref36,ref37}. One method is to use a linear function, illustrated in Figure \ref{fig:curve1.png}a , defined in Equation (5). The value of RUL decreases linearly, with the degradation rate remaining constant. However, during the actual operation of the bearing, the degradation rate of the bearing does not remain constant. Therefore, the experimental results of this labeling method might not be optimal.

\begin{equation}
y(t_i)=-\left(\frac{1}{t_n}*t_{i}\right)+1
\end{equation}
where $t_n$ is the total number of samples, and $t_i$ denotes the sequence number $i$.

And the other employs a piecewise function to predict the degradation trend of bearings. As depicted in Figure \ref{fig:curve1.png}b, the RUL stays consistent up until time $t_j$=0.4 and subsequently diminishes in a linear manner until the bearing reaches its complete failure at time $t_n$. The primary hurdle of this technique is pinpointing the precise onset of the FPT. Additionally, this strategy tends to neglect faint patterns present in initial data, and the rate of degradation isn't linear during the rapid deterioration phase.

\begin{equation}
    y(t_i)=\begin{cases}1&t_i\leq t_j\\\left(\frac{1}{t_j-t_n}\right)*t_i+\left(\frac{t_n}{t_n-t_j}\right)&t_i>t_j\end{cases}
\end{equation}
where $t_n$ is the total number of samples,  $t_i$ denotes the sequence number $i$  and $t_j$ is the FPT.

Given the myriad issues associated with the preceding two methods, it is a common practice to construct  HI curves as labels. An optimal HI exhibits non-linear behavior, maintains a certain degree of monotonicity, and experiences minimal fluctuations within a confined range, with a value domain spanning from 0 to 1. The idea for HI usually involves first transforming the waveform into corresponding features, followed by feature selection to eliminate features with poor monotonicity and trendency. Subsequently, feature dimensionality reduction is carried out, where methods like PCA (principal component analysis), autoencoder, and others can be utilized. 
In this study, a suite of traditional feature extraction methodologies in the time-frequency domain has been utilized. The selected time domain features encompass (root mean square, variance, peak value, peak-to-peak value, skewness, kurtosis,  peak index, margin index and  waveform index), alongside frequency domain attributes such as root mean square frequency and frequency center of gravity. Subsequent to the discrete wavelet packet decomposition, the original signal undergoes a three-tiered wavelet packet decomposition, yielding eight distinct sub-bands. The energy content of each sub-band is then quantified and harnessed as a characteristic feature.  It is observed that vibration data typically manifest in two principal orientations: lateral and longitudinal. From these, a set of 38 distinct features is derived, which are subsequently condensed via PCA to delineate the bearing's state degradation trajectory. This method is referred to as the "feature+PCA" method in this paper, which is also abbreviated as "F+PCA". 

In a similar vein, features can be extracted using an autoencoder. This method is referred to as "feature+AE", which is also abbreviated as "F+AE". The process results in a 1x9 vector, the structure of which is depicted in Table \ref{tab:small-autoencoder}. In the case of directly employing an Autoencoder (AE), the encoder generates a 1x32 vector. Subsequently, a SOM(self-organizing map) network\cite{ref32} is trained using the  vector obtained from the autoencoder. This network, trained through unsupervised learning, facilitates a low-dimensional (typically two-dimensional) discretized representation of the input space of the training samples. Each node on the SOM grid competes to be the closest to the input vector, and over time, the nodes self-organize to form a map where spatial proximity corresponds to feature similarity. Following the training, the best matching node corresponding to each sample is identified. The HI value for each sample can be derived by computing the distance between the sample and its corresponding node. The formula for this can be expressed as:

\begin{equation}    
HI =||x_i-w_i||
\end{equation}
where $x_i$ represents the vector representation  of the $i^{th}$ sample after being processed through the encoder of AE, and $w_i$ denotes the vector representation of the best matching node  corresponding to $x_i$ in the SOM network.

For the VQ-VAE network, the codebook and SOM serve the same clustering function; therefore, we also have an approach called "feature+VQVAE", which is abbreviated as "F+VQVAE". The dimensions of the latent variables produced by the direct use of VQ-VAE and "F+VQVAE" are 32x16 and 9x4, respectively. The structure of these methods are presented in Tables \ref{tab:VQ-VAE} and Tables \ref{tab:small-VQVAE}, respectively. Absolutely, the integration of the clustering process within the neural network framework in VQ-VAE, unlike in SOM, facilitates end-to-end training. This along with its principled regularization through variational inference not only eases its integration into larger neural network architectures but also potentially enhances generalization across varying data scenarios. We use the same formula to represent HI. The formula is as follows:
\begin{equation}    
HI =||x_i-w_i||
\end{equation}
where $x_i$ represents the vector representation  of the $i^{th}$ sample after being processed through the encoder of  VQ-VAE,  and $w_i$ denotes the vector representation of the nearest code corresponding to $x_i$ in the codebook.




\section{Experiments}\label{section:exp}
\subsection{Dataset Introduction}

\begin{figure}
    \centering
    \includegraphics[width=0.75\linewidth]{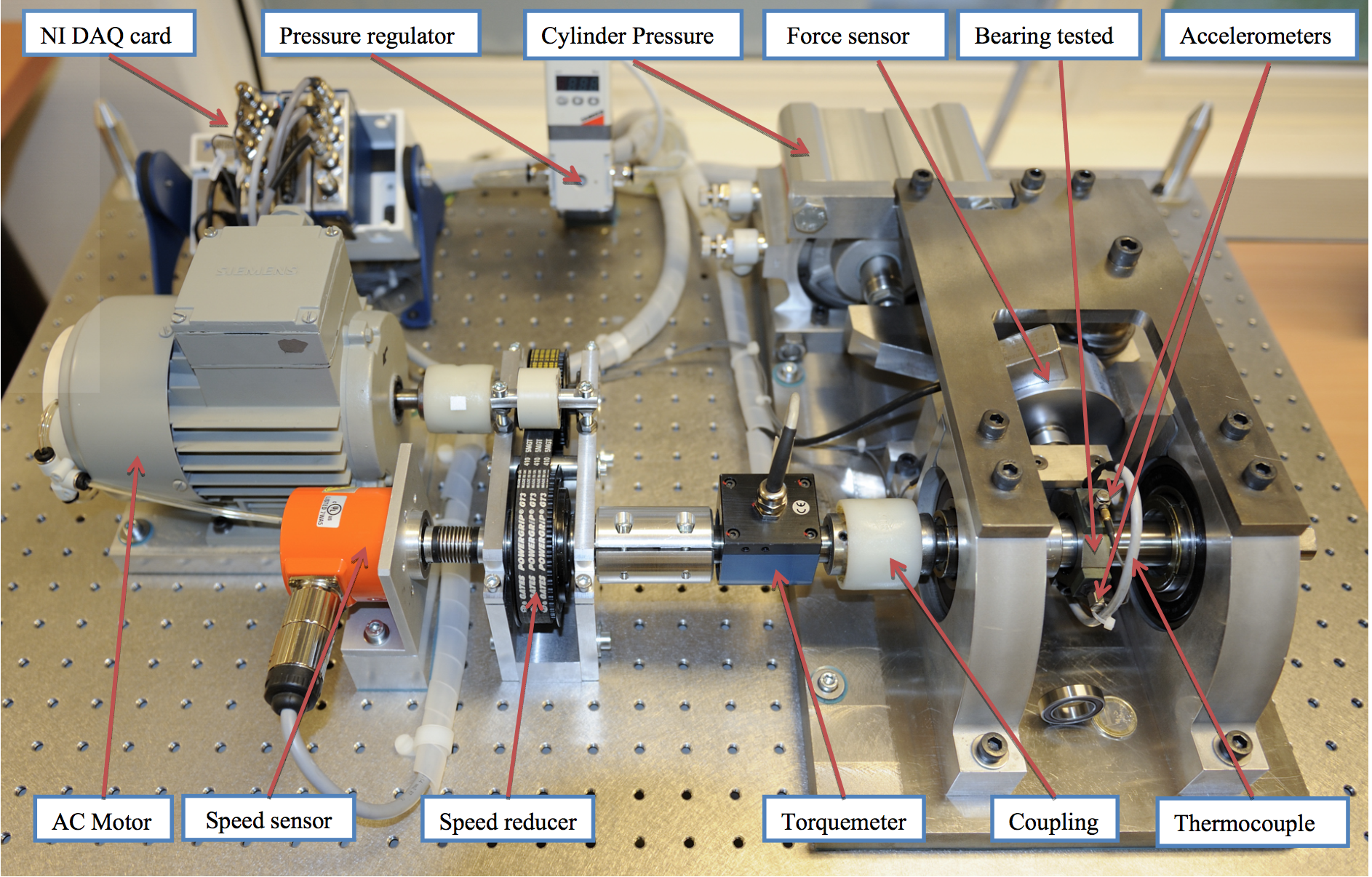}
    \caption{Overview of PRO-NOSTIA}
    \label{fig:dataset}
\end{figure}

Our method was validated using a dataset for bearing performance run-to-failure \cite{ref38}. This dataset, presented at the IEEE PHM2012 Data Challenge, was captured with the PRO-NOSTIA testing setup, as depicted in Figure \ref{fig:dataset}. Data gathering involved two accelerometers, oriented horizontally and vertically. With a sampling rate set at 25.6 kHz, readings were taken every 10 seconds for a duration of 0.1 seconds, resulting in a total of 2560 data points for each instance. Due to safety precautions, the testing was halted if the vibration data amplitude surpassed 20 g. The time at which this amplitude threshold was crossed was considered as the bearing's failure time. Our predictive framework was assessed utilizing the run-to-failure data under two conditions, illustrated in Table \ref{tab:dataset}. 

During the prediction model training, the AdamW optimizer is utilized to minimize the value of the loss function, with a learning rate set at 0.001 and a batch size of 128. The network is trained over 100 epochs, and an early stopping mechanism is employed to prevent overfitting, with the early stopping parameter set at 15.
In the training process of the label model, the AdamW optimizer with a learning rate of 0.001 is utilized, and a batch size of 256 is set. The neural network is trained for 150 epochs with an early stopping parameter set to 20 to prevent overfitting.Results presented in this paper, which encompass both the label and prediction models, are averaged over three independent runs using random seeds 15,  16,  and 25 to ensure consistency and reproducibility. All experiments are conducted on an NVIDIA GeForce 3090 (24GB) GPU, and the models are implemented using the PyTorch framework.

\begin{figure}[htbp]
    \centering
    \begin{tabular}{cc}
        \includegraphics[width=0.45\textwidth]{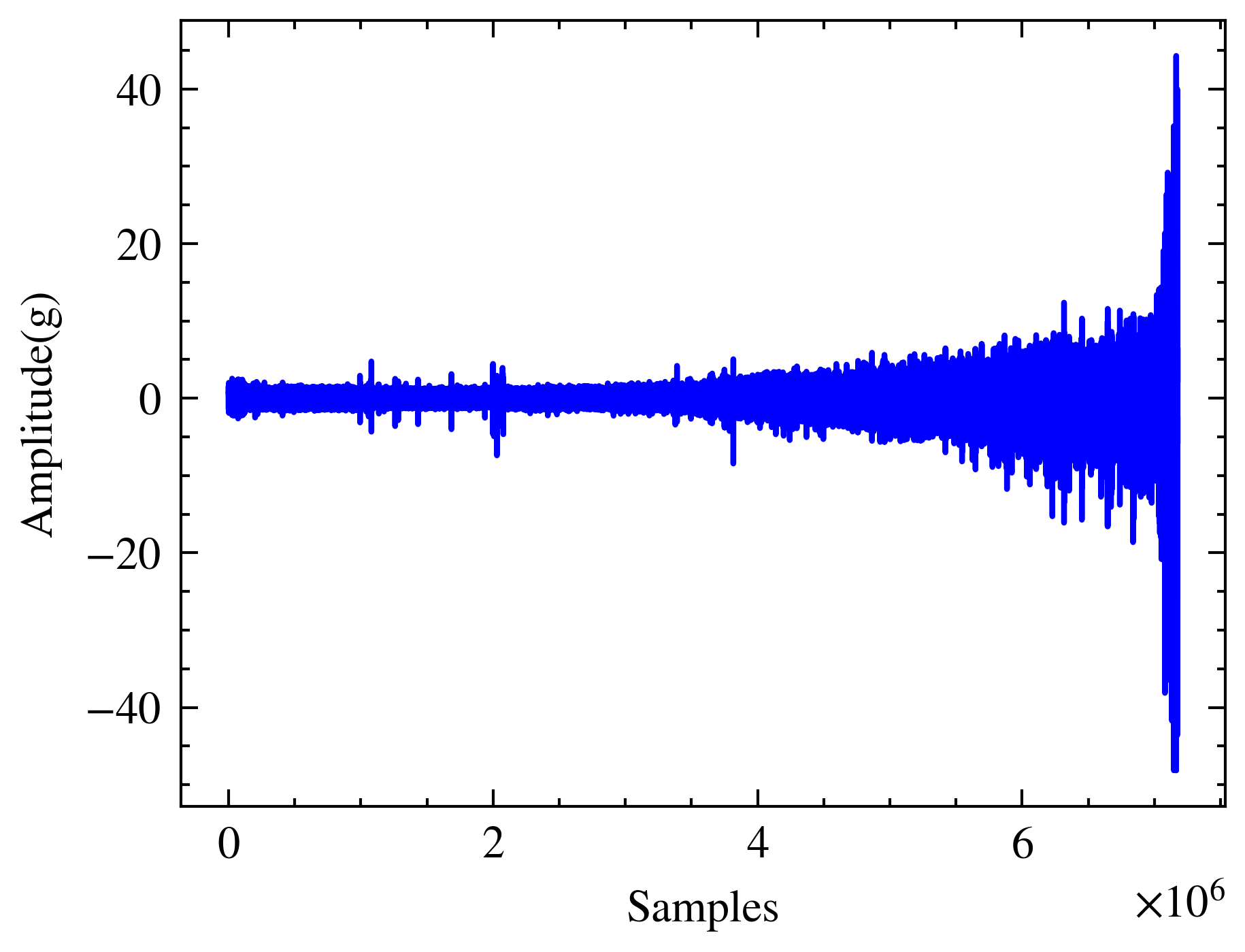} & \includegraphics[width=0.45\textwidth]{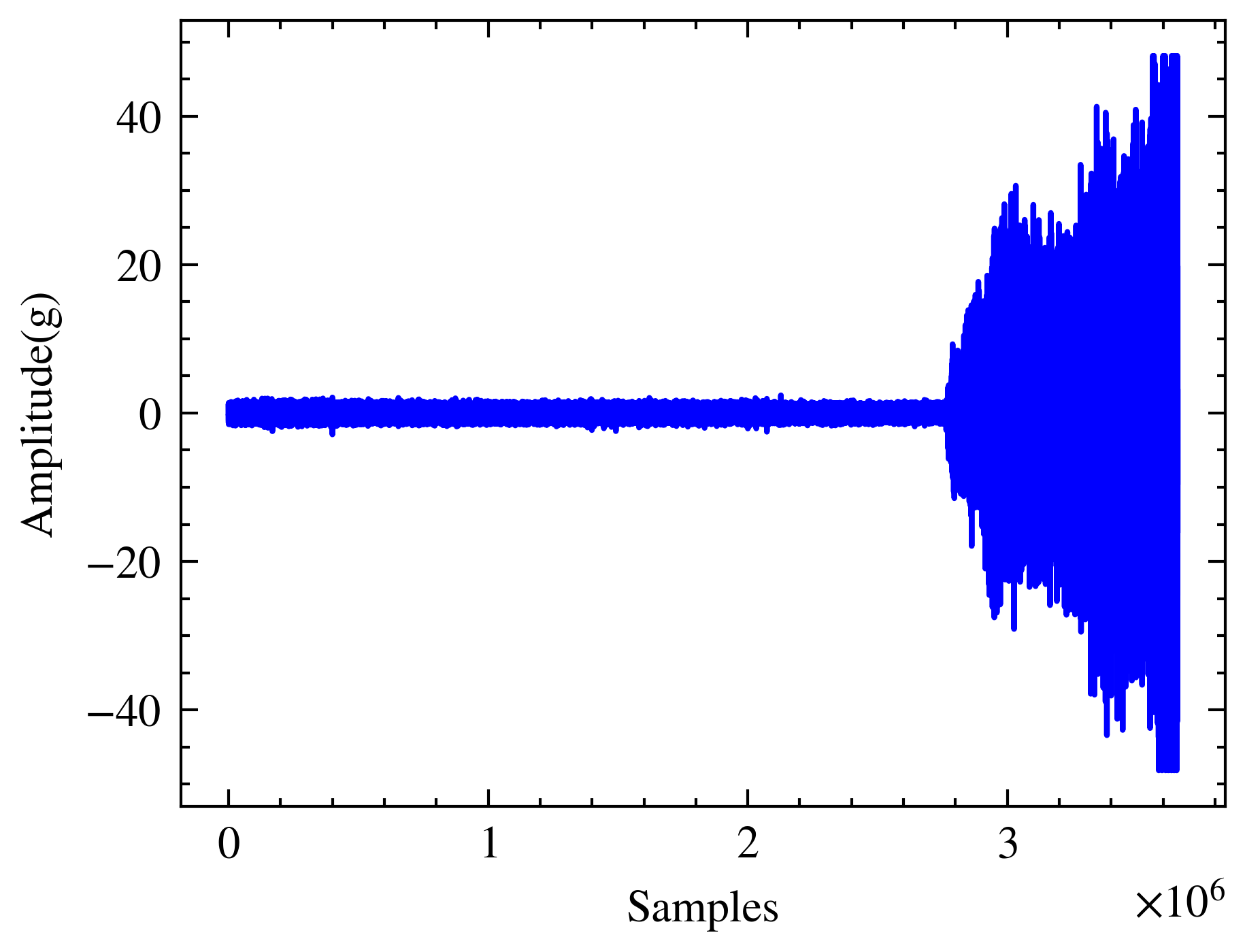} \\
        (a) bearing 1-1 & (b) bearing 1-4 \\
    \end{tabular}
    \caption{The horizontal vibration signals of PHM2012 dataset. (a) bearing 1-1. (b) bearing 1-4.}
    \label{fig:bearing}
\end{figure}

The alternate method is employed to divide the dataset into training and testing sets; that is, selecting one bearing from the seven available, by its identifier, as the test set while using the remaining six bearings for the training set. 
The label model is trained using the data from all seven bearings under operating condition 1. The results and comparisons presented in this paper primarily focus on bearings 1-1 and 1-4 to demonstrate the performance of various label generation methods. The life cycle data is shown in Figure ~\ref{fig:bearing}.

\begin{table} \caption{ Bearing full life under two different working conditions.} \centering \begin{tabularx}{\textwidth}{XXXXX} \toprule \textbf{Operating Conditions} & \textbf{Working Load} & \textbf{Rotating Speed (r/min)} & \textbf{Bearing Number} & \textbf{Actual Life (s)} \\ \midrule \multirow{7}{*}{condition one} & \multirow{7}{*}{4000 N} & \multirow{7}{*}{1800} & Bearing1-1 & 28,030 \\ & & & Bearing1-2 & 8,710 \\ & & & Bearing1-3 & 23,750 \\ & & & Bearing1-4 & 14,280 \\ & & & Bearing1-5 & 24,630 \\ & & & Bearing1-6 & 24,480 \\ & & & Bearing1-7 & 22,590 \\ \midrule \multirow{5}{*}{condition two} & \multirow{5}{*}{4200 N} & \multirow{5}{*}{1650} & Bearing2-1 & 9,110 \\ & & & Bearing2-2 & 7,970 \\ & & & Bearing2-3 & 19,550 \\ & & & Bearing2-4 & 7,510 \\ & & & Bearing2-5 & 23,110 \\ \bottomrule \end{tabularx}  \label{tab:dataset} \end{table}

\subsection{Metrics}
We have chosen the most commonly used RMSE (root mean square error) metric to evaluate the difference between the predicted values and the labels. It represents the square root of the quadratic mean of the differences between predicted values and observed values, as the Equation (9) shows.
\begin{equation}RMSE=\sqrt{\frac{1}{N}\sum_{i=1}^{N}\left(\hat x_i-x_{i}\right)^{2}}\end{equation}
where $\hat x_i$ denotes actual RUL of the $i^{th}$ predictive object, $x_i$ signifies the predicted RUL of the $i^{th}$ predictive object and $N$ denotes the number of the samples. 

To ensure the validity of the prediction results, we select the score metric proposed by the PHM2012 official challenge\cite{ref38}, and its formula are as follows:

\begin{equation}
\begin{aligned}
E_{i}&=\hat x_{i}-x_{i}\\
A_i&=\begin{cases}
\exp^{-(\frac{E_{i}}{13})}
&\text{ if }E_{i}\leq0\\
\exp^{(\frac{E_{i}}{10})}
&\text{ if }E_{i}>0 
\end{cases} \\
Score&=\frac{1}{N}\sum_{i=1}^{N}(A_{i}) 
\end{aligned}
\end{equation}
where $\hat x_i$ denotes actual RUL of the $i^{th}$ predictive object, $x_i$ signifies the predicted RUL of the $i^{th}$ predictive object, and $N$ denotes the total number of samples.

In order to conduct a quantitative analysis of the labels, this paper selects monotonicity and trendency indicators from commonly used health indicator evaluation metrics. Monotonicity measures the monotonous variation trend of the HI curve, while trendiness reflects the correlation between degradation stages.

\begin{equation}
\begin{aligned}
mon&=\left|\frac{Num~of~\delta x>0}{N-1}-\frac{Num~of~\delta x<0}{N-1}\right|\\\\
corr&=\frac{\left|\sum_{i=1}^T\left(x_i-\bar{x}\right)(l_i-\bar{l})\right|}{\sqrt{\sum_{i=1}^T\left(x_i-\bar{x}\right)^2\sum_{i=1}^T\left(l_i-\bar{l}\right)^2}}
\end{aligned}
\end{equation}
where $\delta x$ denotes the difference between two adjacent points on the HI curve, $x_i$ represents the predicted RUL for the $i_{th}$ sample, and $l_i$ is the index number corresponding to that sample. Furthermore, $\bar{x}$ is the mean of the predicted RUL values, $\bar{l}$ is the mean of the sample indices, and $N$ represents the total number of samples.

The aforementioned indicators, monotonicity and trendiness, may not adequately reflect the fluctuation characteristics of the data. Therefore, we propose two metrics: mean absolute distance between two adjacent points and mean variance between windows, to provide a more nuanced analysis of the data.

The mean absolute distance (MAD) method is straightforward, involving the computation of the absolute distance between every pair of adjacent points, followed by summing these distances and taking the average. The core idea behind this method hinges on the notion that the HI represents a linear progression from 1 to 0. The cumulative movement distance of all points remains constant;  Conversely, if the points exhibit more significant fluctuations, implying a more pronounced up and down movement, the movement distance increases, suggesting a less accurate HI.
Its equations are shown as follows:

\begin{equation}
mad=\frac1{n-1}\sum_{i=2}^n|x_i-x_{i-1}|
\end{equation}
where  $x_i$ denotes the predicted RUL of the $i^{th}$ predictive object and $n$ represents the total number of samples.

The primary concept of the variance method revolves around segregating continuous data points into designated windows, followed by the computation of the variance of data within each window, and culminating in the averaging of the variances across all windows. The windows are designed to overlap, and after each computation, the window is shifted forward by one data point. This overlapping and incremental shifting approach facilitates a continuous and smooth analysis across the data stream, ensuring that the evaluation captures the data's behavior in a comprehensive and nuanced manner. 
Its equations are shown as follows:
\begin{equation}
\begin{aligned}
\mu_i&=\frac1n\sum_{k=1}^nx_{ik} \\
\sigma_i&=\frac{1}{k}\sum_{j=0}^{k-1}(x_{i+j}-\mu_i)^2 \\
mv&=\frac1{N-k+1}\sum_{i=1}^{N-k+1}\sigma_i
\end{aligned}
\end{equation}
where $k$ denote the length of each window, $x_{i+j}$ represent the predicted value of the $j^{th}$ predictive object in the $i^{th}$ window, $N$ signify the total number of data points in the sample, $\sigma_i$ denote the variance of the window starting from the $i^{th}$ data point, and $\mu_i$ represent the mean of the window starting from the $i^{th}$ data point.

\subsection{Results Analysis}
\subsubsection{Label Model Analysis}
In this section, we will conduct a comprehensive comparison of different label creation methods. For the selection of FPT points in segmented labels, we have employed method as proposed by Zhang et al.\cite{ref39}. The method is based on the $3\sigma$ technique, as illustrated in Table ~\ref{segment}.

For HI, apart from the previously mentioned "feature+PCA" ,  "feature+AE"  and "feature+VQVAE" methods, we have also introduced three additional methods, namely RMS\cite{ref23}, direct AE  and  direct VQ-VAE, for comparison. All these HI curves are smoothed using the Savitzky-Golay filter with a window length of 21.

\begin{table}
    \caption{ FPT of PHM2012 dataset under condition 1. \label{segment} }
    \begin{tabularx}{\textwidth}{CCC}
        \toprule
        \textbf{Bearings}  & \textbf{FPT\cite{ref39}}   &\textbf{Actual Life(s)}  \\
        \midrule
        B1   & 11420 & 28030 \\
        B2   & 8220 & 8710 \\
        B3   & 9600 & 23750 \\
        B4  & 10180 & 14280 \\
        B5  & 24070 & 24630 \\
        B6  & 16270 & 24480 \\
        B7  & 22040 & 22590 \\
        \bottomrule
        \end{tabularx}
\end{table}

\begin{figure}[H]
    \centering
    \begin{tabular}{cc}
        \includegraphics[width=0.5\textwidth]{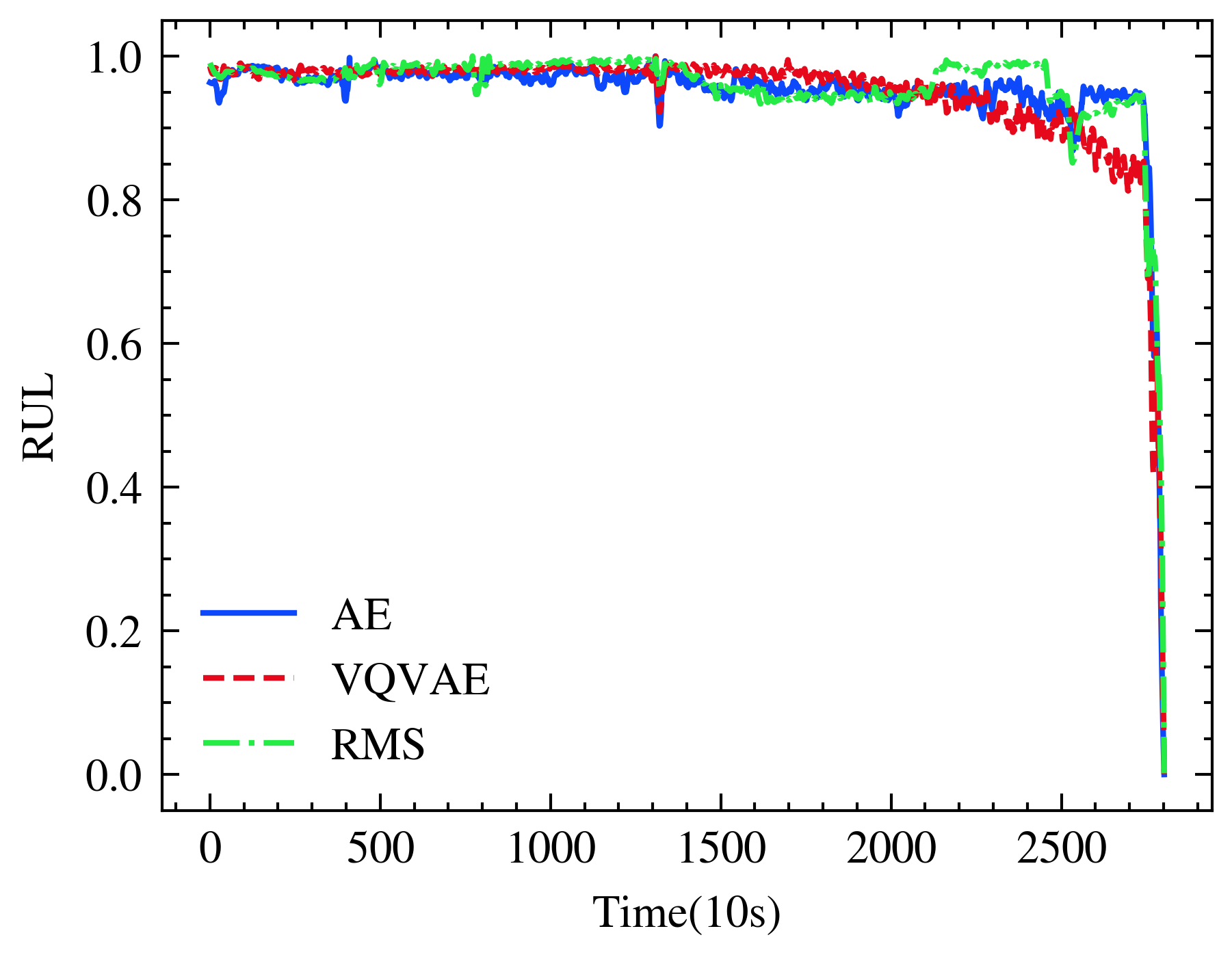} & \includegraphics[width=0.5\textwidth]{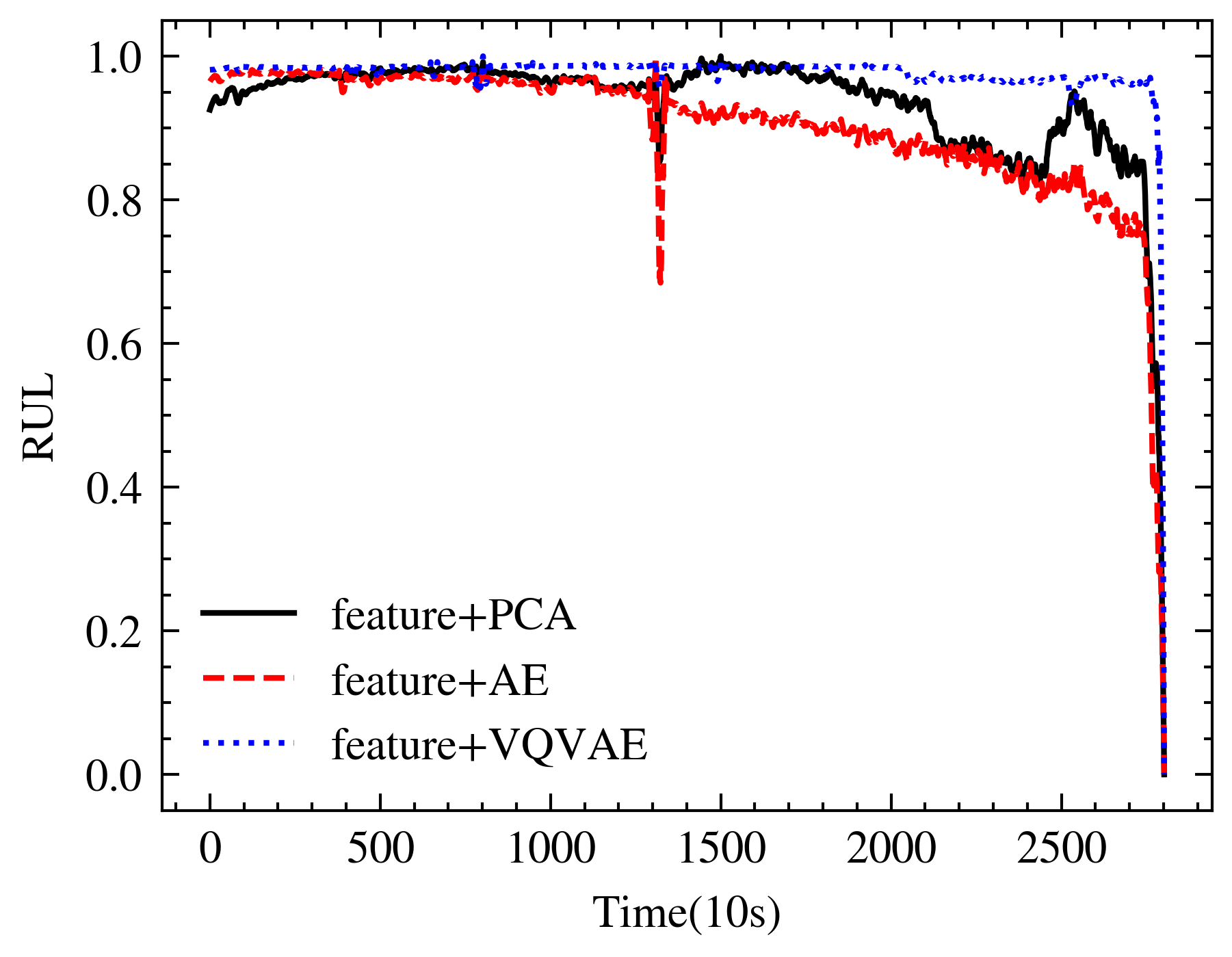} \\
        (a)  & (b)  \\
    \end{tabular}
    \caption{Labels of bearing 1-1 constructed based on six different methods. (a)Direct generation. (b) Feature Selection and generation. }
    \label{fig:label1-1}
\end{figure}

\begin{figure}[H]
    \centering
    \begin{tabular}{cc}
        \includegraphics[width=0.5\textwidth]{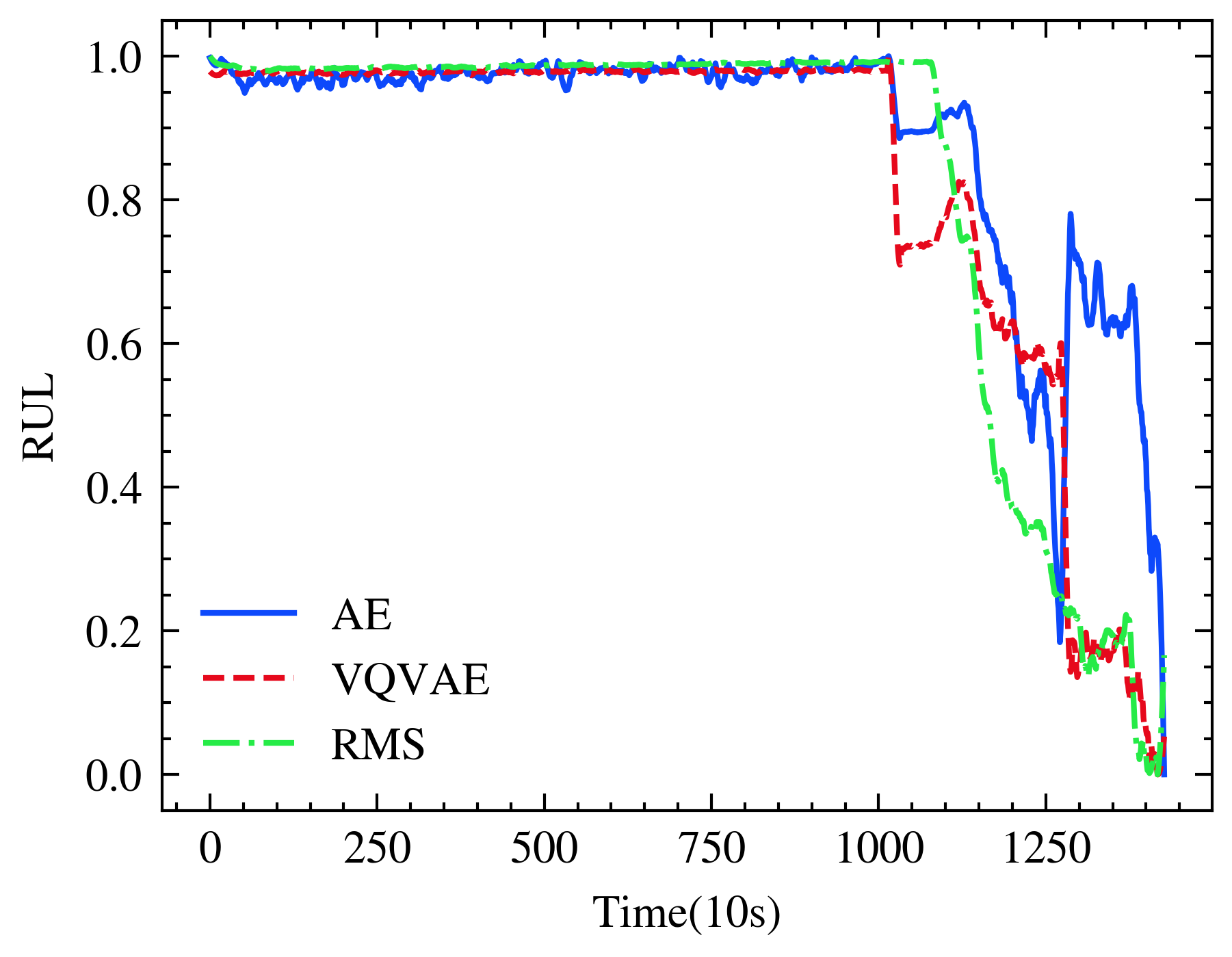} & \includegraphics[width=0.5\textwidth]{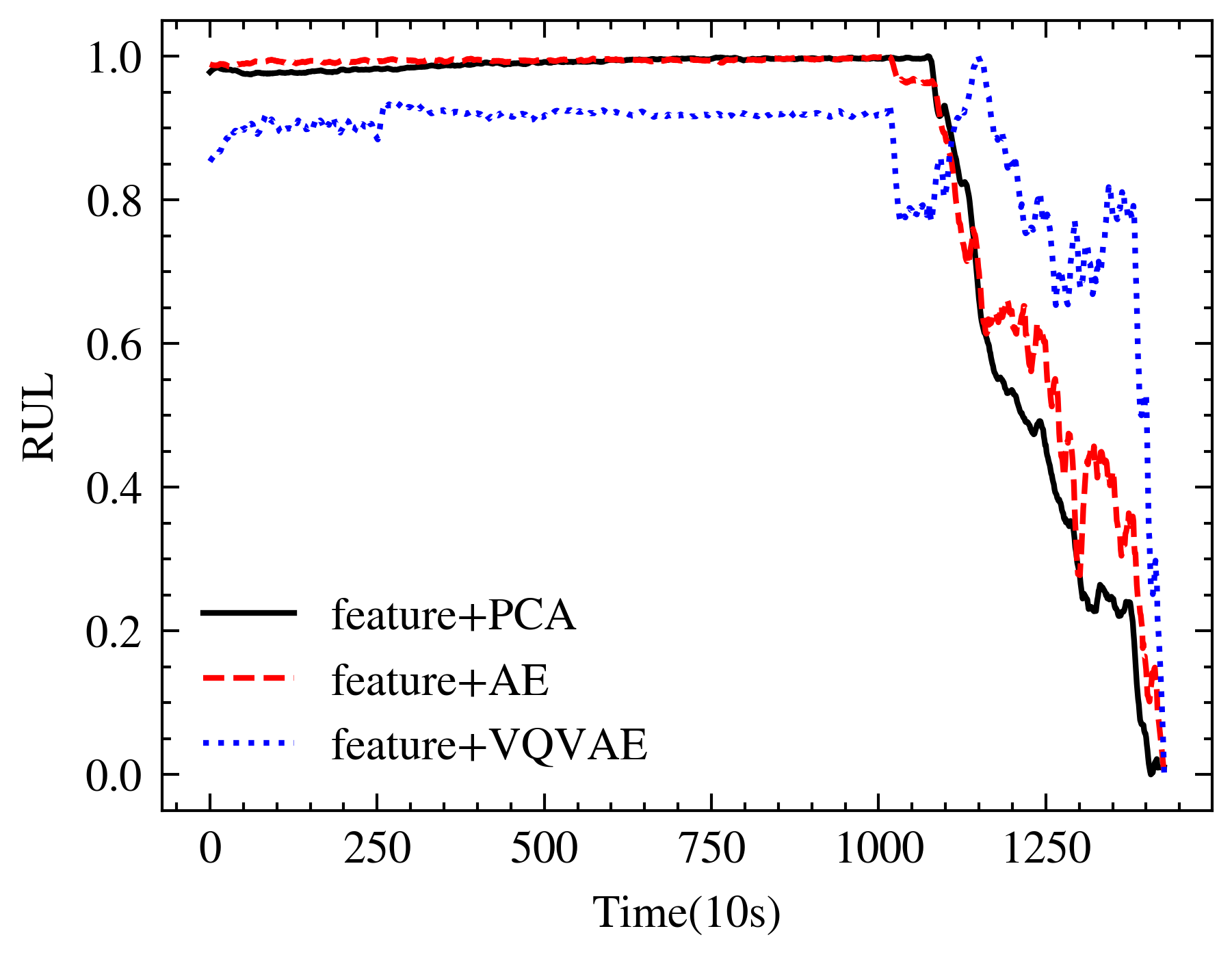} \\
        (a)  & (b)  \\
    \end{tabular}
    \caption{Labels of bearing 1-4 constructed based on six different methods. (a)Direct generation. (b) Feature Selection and generation.}
    \label{fig:label1-4}
\end{figure}
Figure \ref{fig:label1-1} shows displays six distinct label creation methods of bearing1-1. In sub-figure (a), it is observed that the prediction performance of these three methods remains relatively stable over most of the time intervals. However, post t=2200, a certain level of fluctuation is evident in the two methods other than VQ-VAE. In sub-figure (b), the "feature+ae" method exhibits a trough at t=1250, while the "feature+VQVAE" method maintains smoothness throughout the entire process.

In Figure \ref{fig:label1-4}, the decline for bearing1-4 is markedly more severe compared to Figure \ref{fig:label1-1}.The two non-neural network methods demonstrate fairly good performance. Observed in Figure \ref{fig:label1-4} (a) is a heavy trough at t=1250 for the AE method, while the other two methods remain relatively smooth. In Figure \ref{fig:label1-4}(b), the "feature+PCA" method maintains smoothness throughout. The initial value prediction of "feature+VQVAE" exhibits a certain degree of offset. 

The test results of these label creation methods on bearings 1 to 7 are shown in Table \ref{tab:label-avg}. The "feature+VQVAE" method achieves the lowest MAD, while the AE method achieves the lowest MV value, with the MV value of VQ-VAE not differing significantly from it. The RMS method, which has the highest monotonicity and tendency, performs poorly on both MAD and MV, indicating that these two traditional metrics cannot accurately measure the fluctuation of the curves.
\begin{table}[!htbp]
    \centering
    \caption{Comparative analysis of the average results of label models for bearings 1 through 7.}
    \begin{tabularx}{\textwidth}{CCCCCCC}
        \toprule
        \textbf{Metrics} & \textbf{AE} & \textbf{F+VQVAE} & \textbf{VQVAE(Our)} & \textbf{F+AE} & \textbf{RMS} & \textbf{PCA} \\ 
        \midrule
        Mon & 0.0310  & 0.0203  & 0.0277  & 0.0176  & \textbf{0.0438}  & 0.0379  \\ 
        Tend & -0.309  & -0.292  & -0.318  & -0.330  & \textbf{-0.408}  & -0.247  \\ 
        MAD & 0.0293  & \textbf{0.0215}  & 0.0310  & 0.0373  & 0.0336  & 0.0479  \\ 
        MV & \textbf{0.00103}  & 0.00135  & 0.00108  & 0.00205  & 0.00388  & 0.00616 \\ 
        \bottomrule
    \end{tabularx}
    \label{tab:label-avg}
\end{table}

\subsubsection{Prediction Model Analysis}
\begin{figure}[H]
    \begin{adjustwidth}{-\extralength}{0cm}
    \begin{tabularx}{\linewidth}{CCC}
        \includegraphics[width=\linewidth]{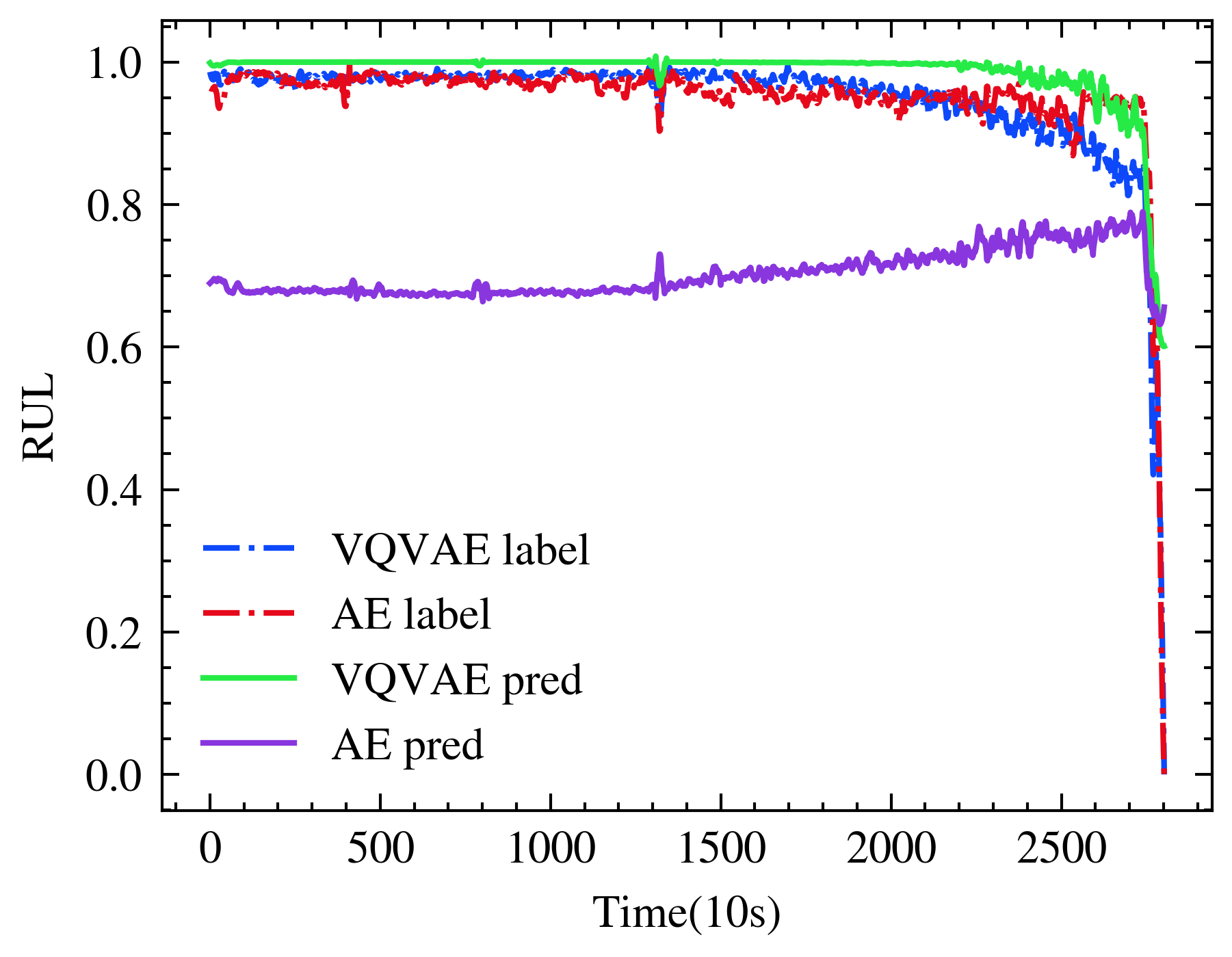} & \includegraphics[width=\linewidth]{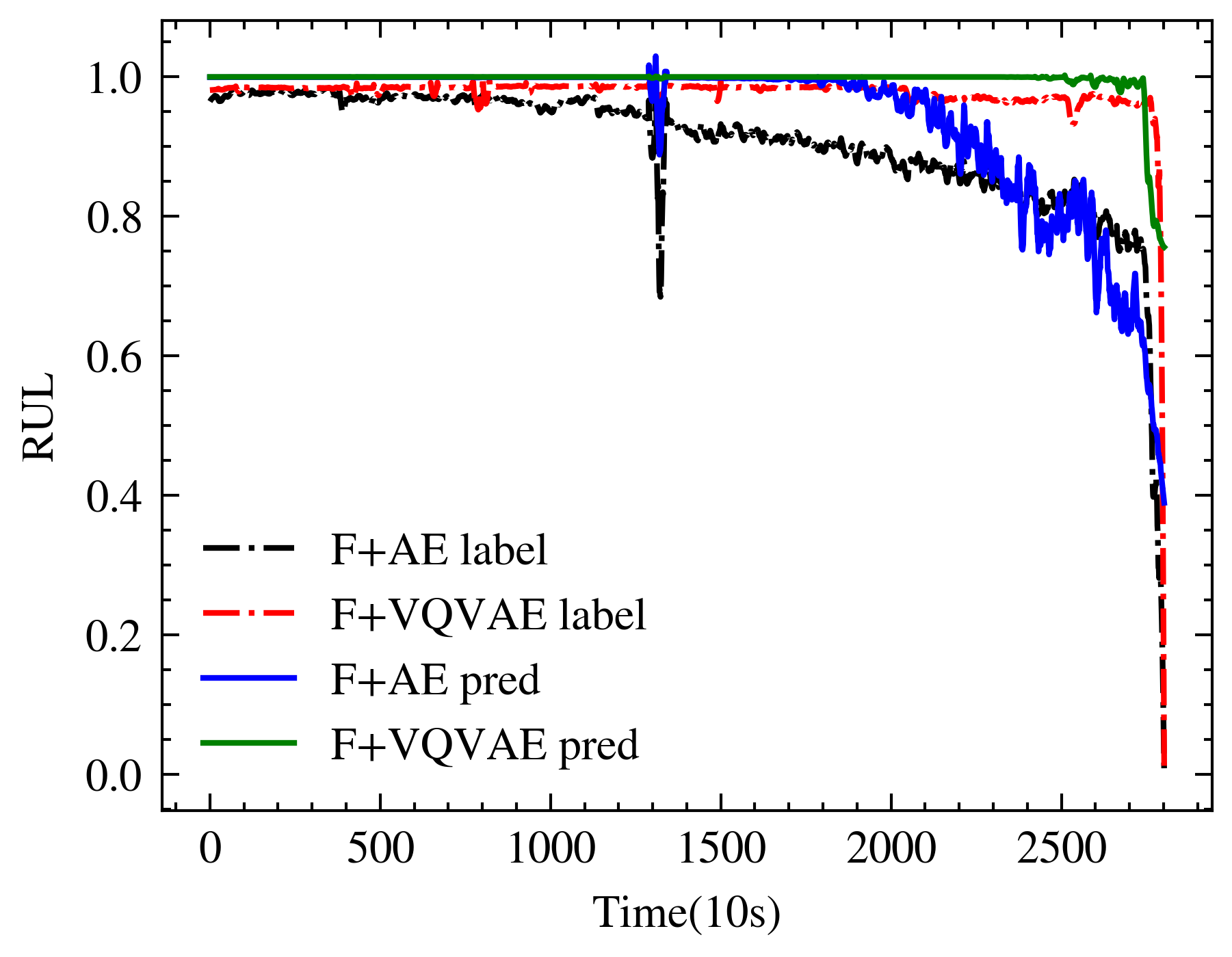}  & \includegraphics[width=\linewidth]{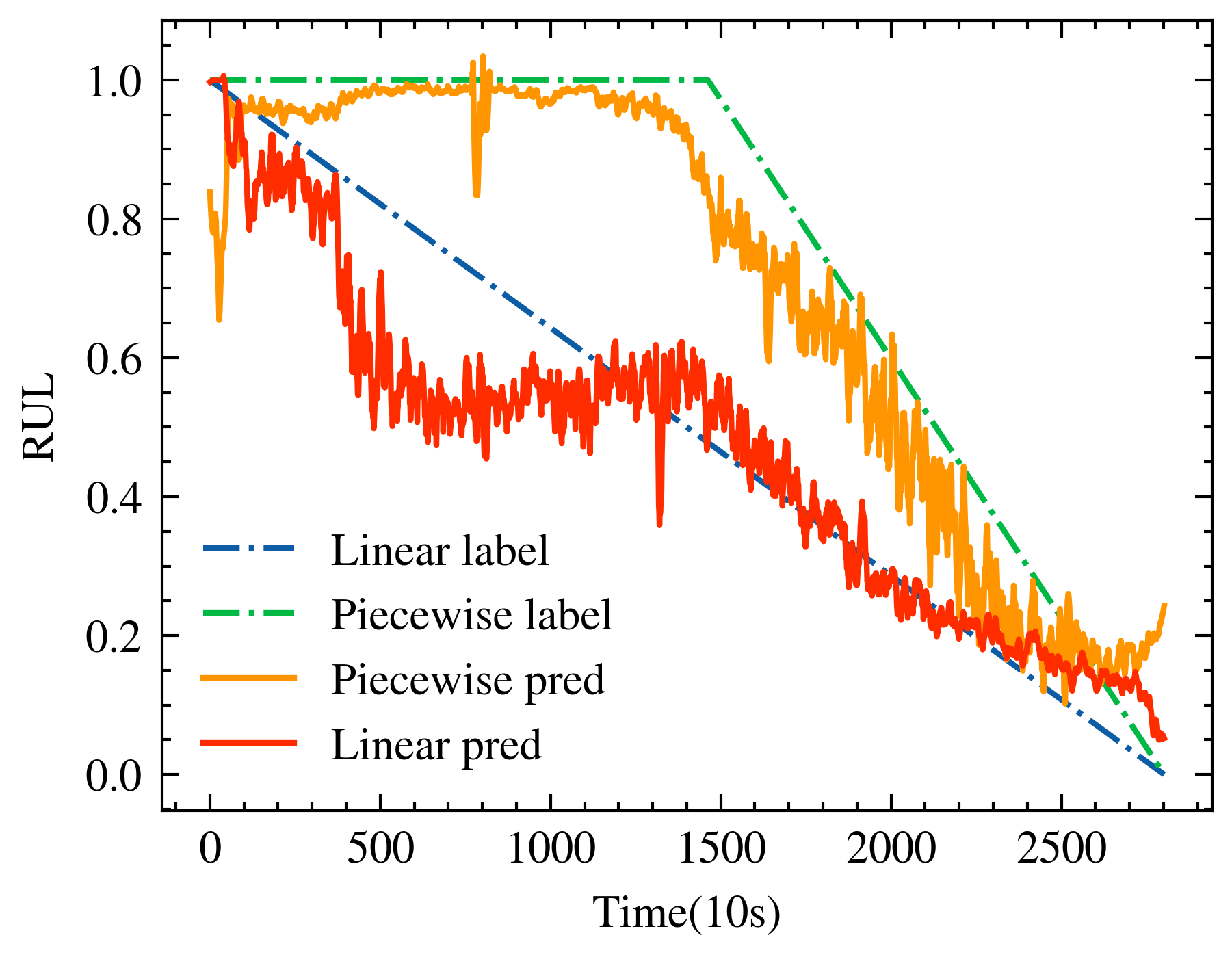}\\
        (a)  & (b) & (c)\\
    \end{tabularx}
    \caption{Prediction of bearing 1-1 trained by six different methods. (a)Direct generation. (b)Feature selection and generation. (c) Tradition method.}
    \label{fig:pred1-1}
    \end{adjustwidth}
\end{figure}

\begin{figure}[H]
    \begin{adjustwidth}{-\extralength}{0cm}
    \begin{tabularx}{\linewidth}{CCC}
        \includegraphics[width=0.40\textwidth]{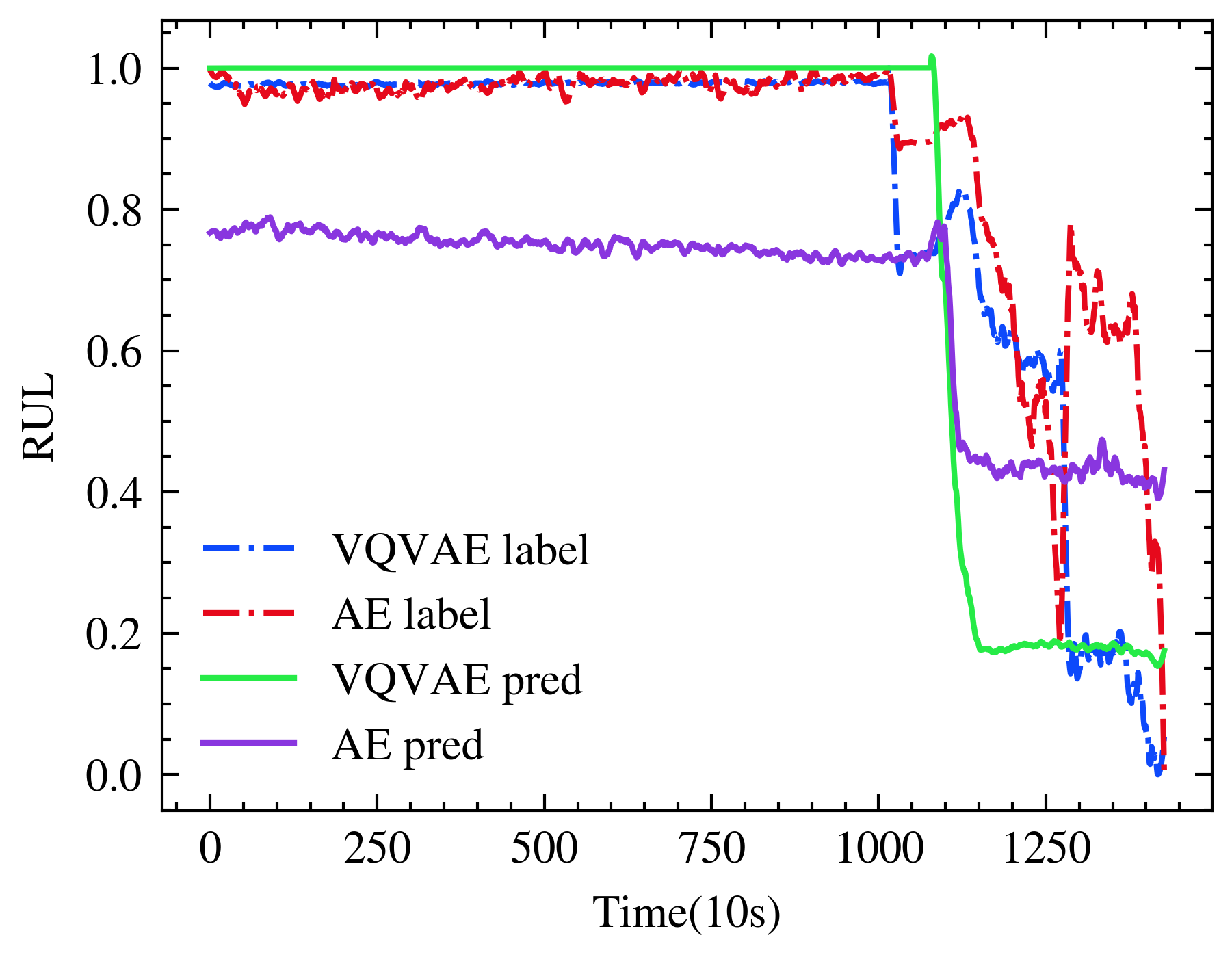} & \includegraphics[width=0.40\textwidth]{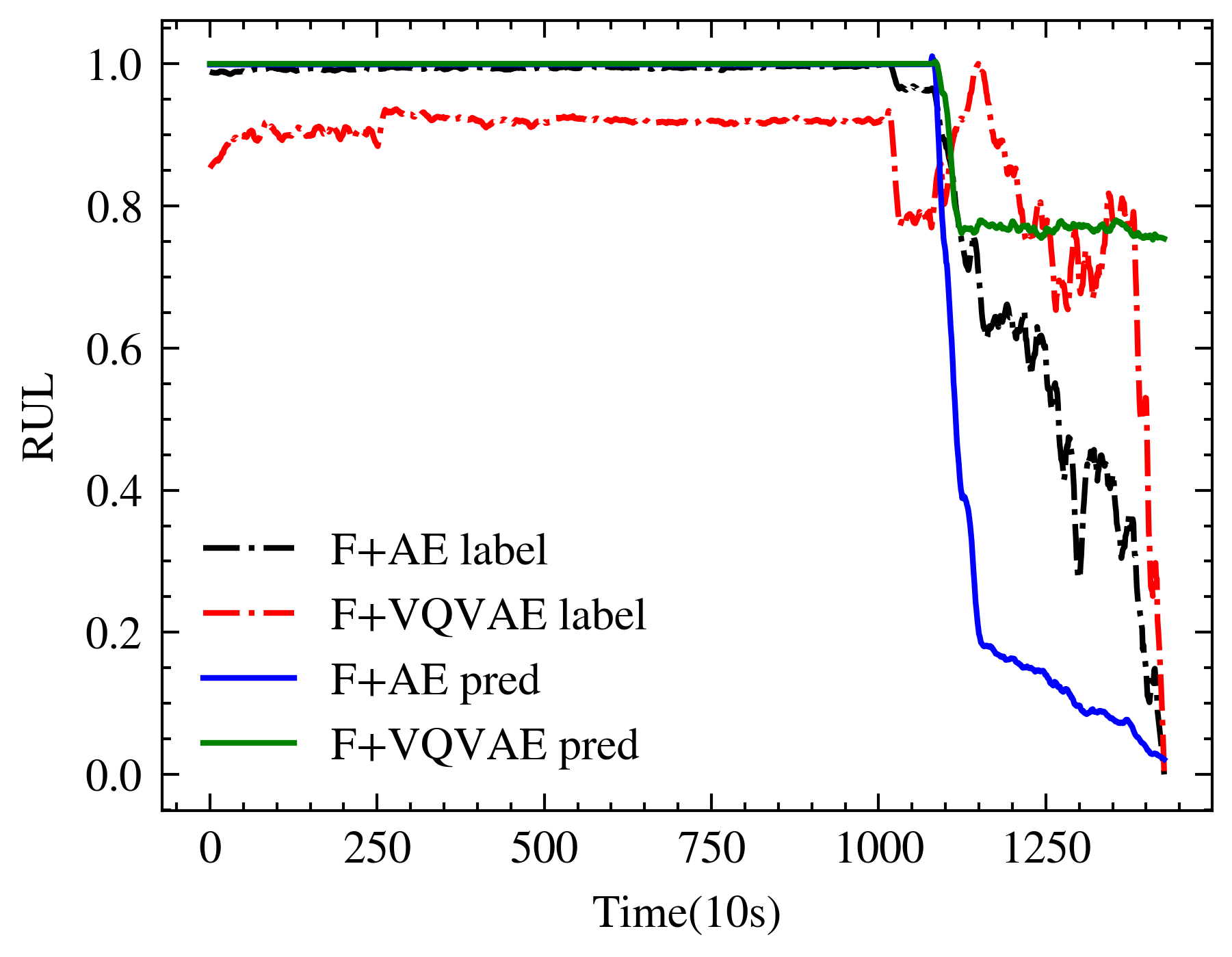} & \includegraphics[width=0.40\textwidth]{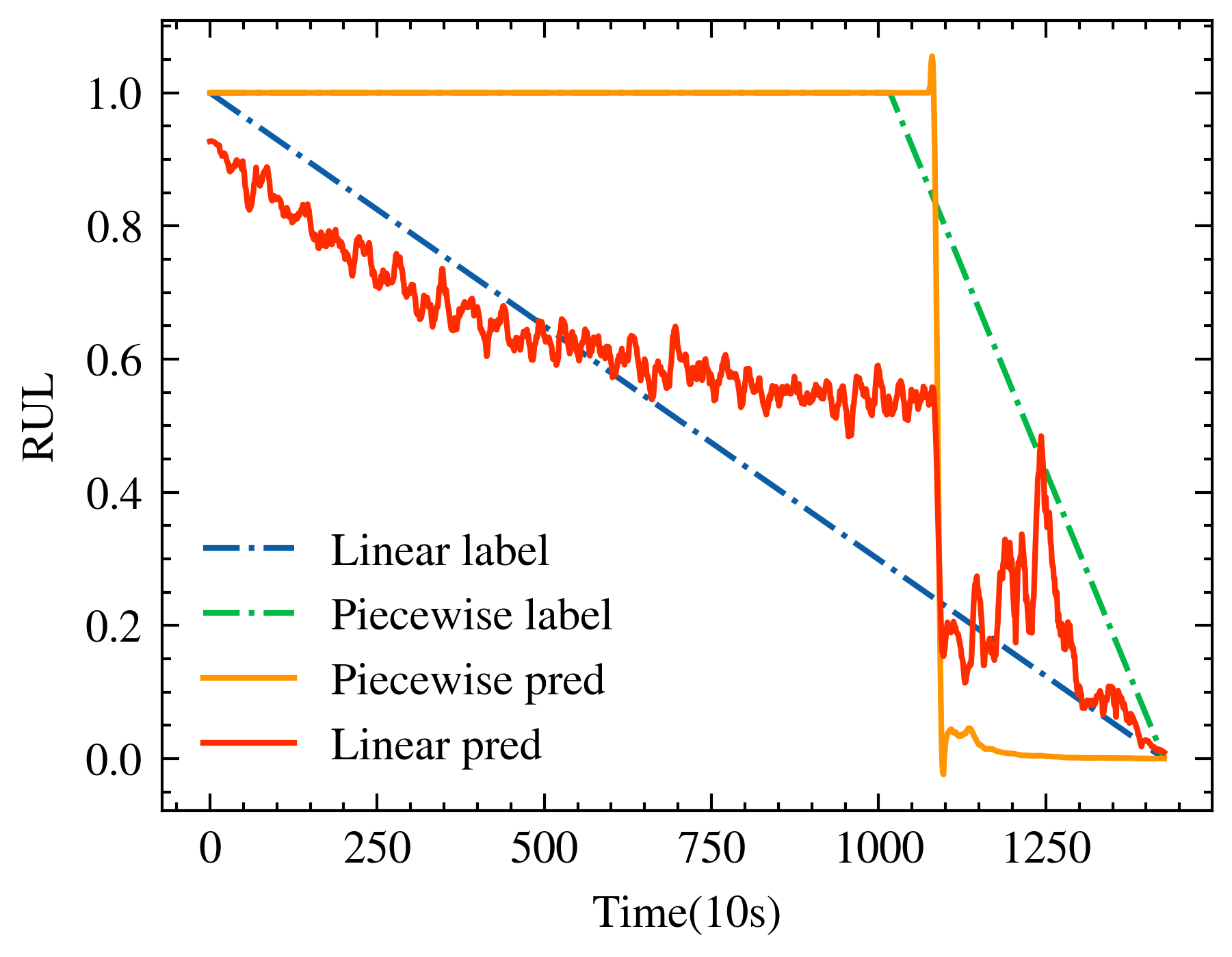}\\
        (a)  & (b)  & (c)\\
    \end{tabularx}
    \caption{Prediction of bearing 1-4 trained by six different methods. (a)Direct generation. (b)Feature selection and generation. (c) Tradition method.}
    \label{fig:pred1-4}
    \end{adjustwidth}
\end{figure}

In this section, the ASTCN model is employed as the prediction model, and the alternating training method is utilized to test different label methods on bearings 1 to 7. In addition to the labels mentioned in the previous subsection, two commonly used labels, linear and segmented, are also tested. The test results are illustrated in Figures \ref{fig:pred1-1} and \ref{fig:pred1-4}, and Table \ref{tab:pred1-7}.

Figure \ref{fig:pred1-1}a displays the prediction results for bearing 1-1, where VQ-VAE method lags in reflecting the degradation trend in the latter part, failing to accurately capture the deteriorating trajectory. In Figure \ref{fig:pred1-1}b, "F+VQVAE" and VQ-VAE methods exhibit very similar patterns, with the only relatively accurate depiction coming from the "F+AE" method, albeit with a degradation trend differing from that of the labels.

Figure \ref{fig:pred1-4}, showcasing the predictions for bearing 1-4, presents a larger discrepancy in the outcomes of different labeling methods, thus better reflecting the variances among different labels. Mirroring the findings in Figure \ref{fig:pred1-1}, VQ-VAE method again exhibits aberrant performance in the latter half, and the result of "F+VQVAE" method closely resembles VQ-VAE method. The prediction of AE method demonstrates an initial value deviation in both figures; however, upon inspecting its prediction results for other bearings in condition 1, no such initial misjudgment as seen in bearing1-1 and bearing1-4 was found. As for the linear and piecewise methods depicted in Figures \ref{fig:pred1-1}c and \ref{fig:pred1-4}c, their performances are comparatively dismal.

Two intriguing conclusions emerge from these figures: "F+AE" method performs well in both, and its predictions even more accurately reflect the degradation trend of the bearings than the labels do, indicating that to some extent, the prediction model has transcended the constraints posed by the labels. The FPT points of bearing1-4 in the labels are all situated earlier than those predicted by the ASTCN model. In conjunction with the RMS and "feature + PCA" non-neural network labels from Figure \ref{fig:label1-4}, it can be posited that these models have overfit to a certain degree.

As shown in Table \ref{tab:pred1-7}, the ASTCN model trained with VQ-VAE labels achieved the optimal performance on MAD and MV metrics, and the RMSE, MAE, and Score values only slightly differed from the optimal "feature+VQVAE" method. This indicates that the models trained with VQ-VAE are able to make more accurate predictions regarding the remaining useful life of bearings. The models trained with piecewise labels also performed well on MAD and MV, but had very high RMSE and MAE values, as shown in Table \ref{tab:preds-all}, being significantly affected by the outliers in bearing 1-3. This suggests that the piecewise labels can be too rigid at times, deviating significantly from the actual predicted values. The ASTCN model trained with "Feature+AE" methods, which performed well in Figures \ref{fig:pred1-1} and \ref{fig:pred1-4}, exhibited very average performance in metrics, likely due to the inaccuracies in the labels.

\begin{table}[!ht]
    \centering
    \caption{Comparative analysis of the average results of prediction models for bearings 1 through 7.}
    \begin{tabularx}{\textwidth}{CCCCCC}
     \toprule
        \textbf{Metrics} & \textbf{RMSE} & \textbf{MAE} & \textbf{MAD} & \textbf{MV} & \textbf{Score} \\ 
        \midrule
        Linear & 0.219  & 0.172  & 0.1002  & 0.00884  & 32.0  \\ 
        Piecewise & 0.210  & 0.144  & 0.0287  & 0.00236  & 27.2  \\ 
        PCA & 0.214  & 0.172  & 0.0566  & 0.00643  & 30.3  \\ 
        AE & 0.156  & 0.142  & 0.0575  & 0.00437  & 26.8  \\ 
        F+AE & 0.128  & 0.097  & 0.0468  & 0.00484  & 17.9  \\ 
        F+VQVAE & \textbf{0.077}  & \textbf{0.060}  & 0.0377  & 0.00348  & \textbf{11.3} \\ 
        VQVAE(Our) & 0.086  & 0.063  & \textbf{0.0275}  & \textbf{0.00122}  & 11.4  \\ 
        \bottomrule
    \end{tabularx}
    
    \label{tab:pred1-7}
\end{table}

\subsubsection{Generalization Test}
\begin{figure}[H]
    \begin{adjustwidth}{-\extralength}{0cm}
    \begin{tabularx}{\linewidth}{CCC}
        \includegraphics[width=\linewidth]{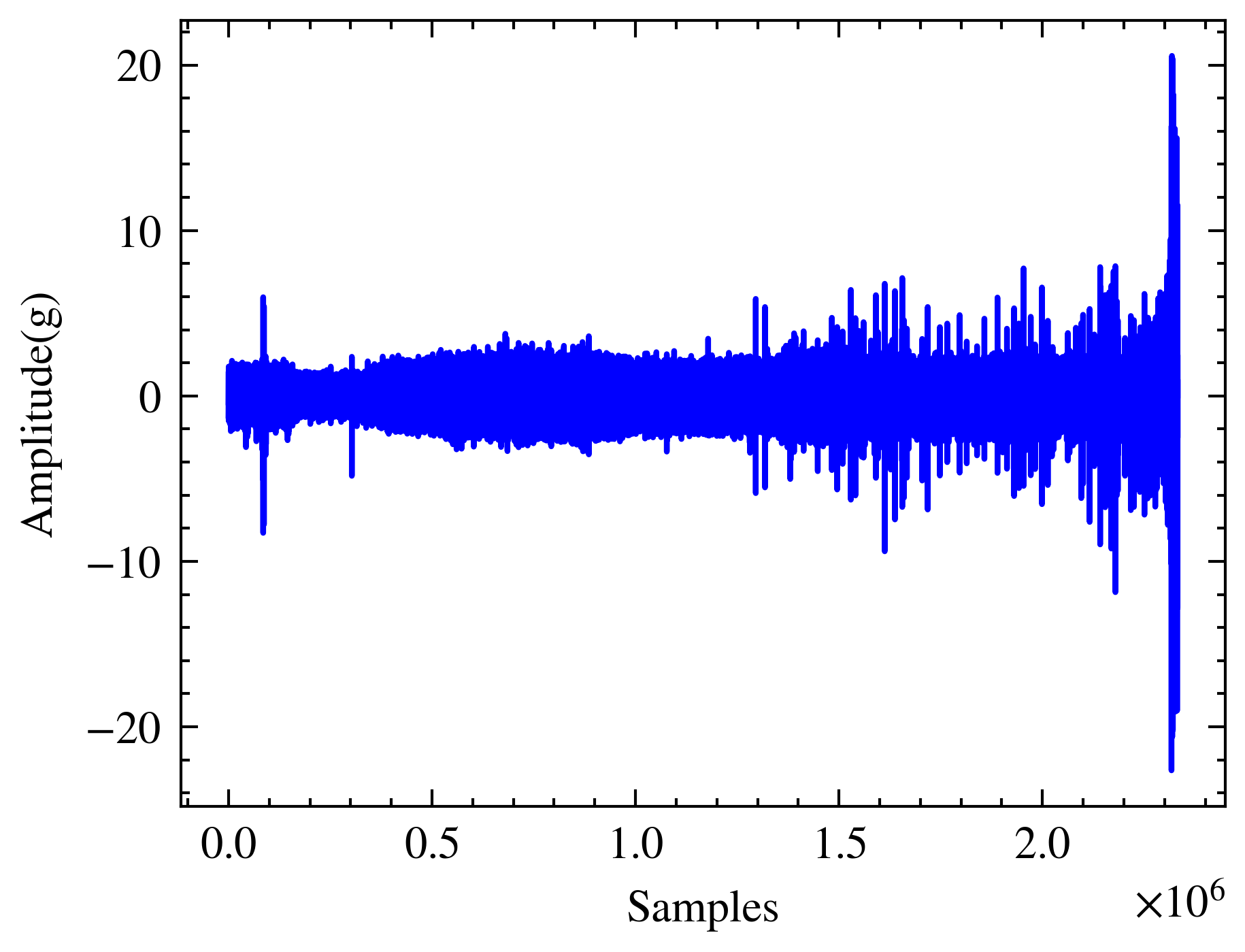} & \includegraphics[width=\linewidth]{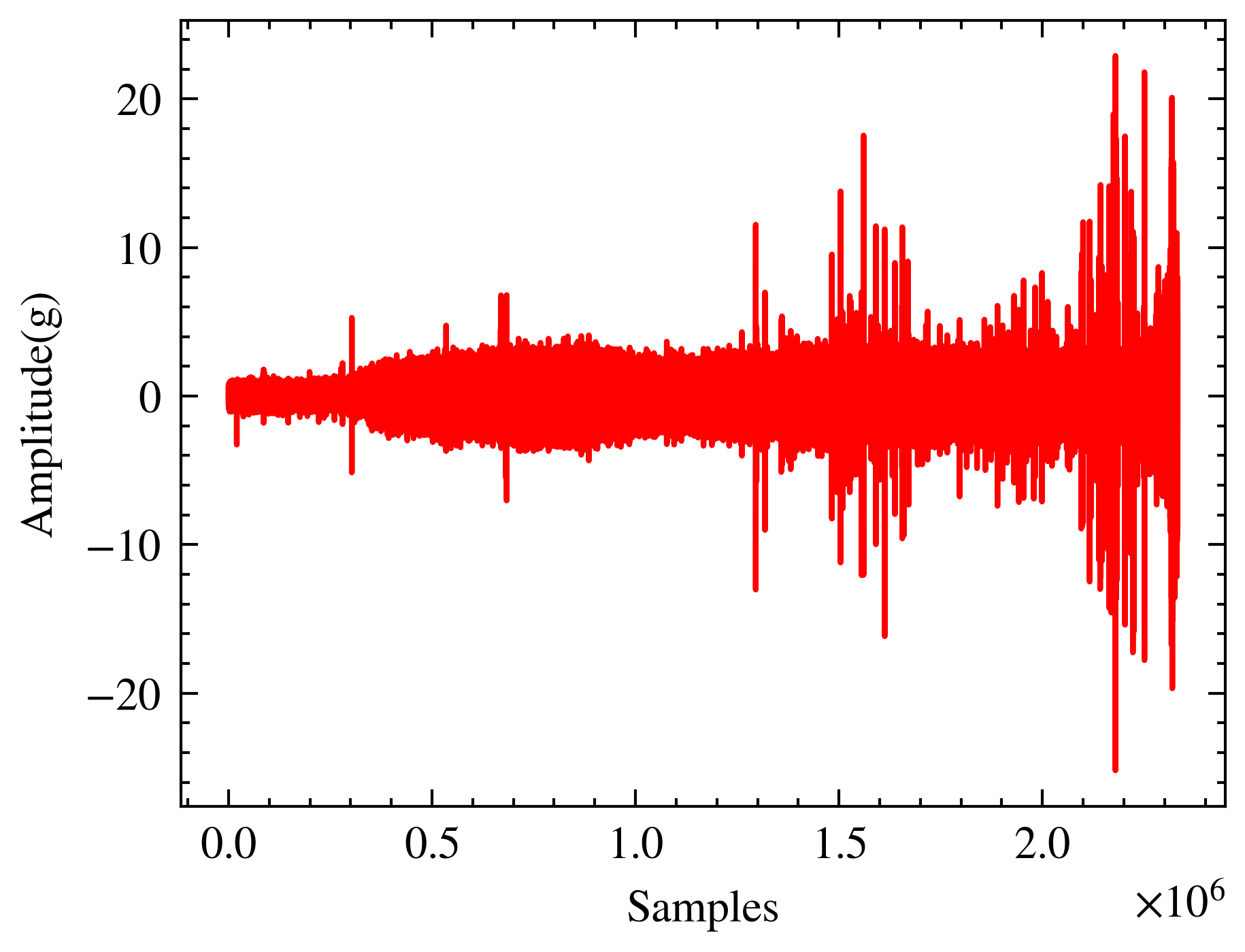} &
        \includegraphics[width=\linewidth]{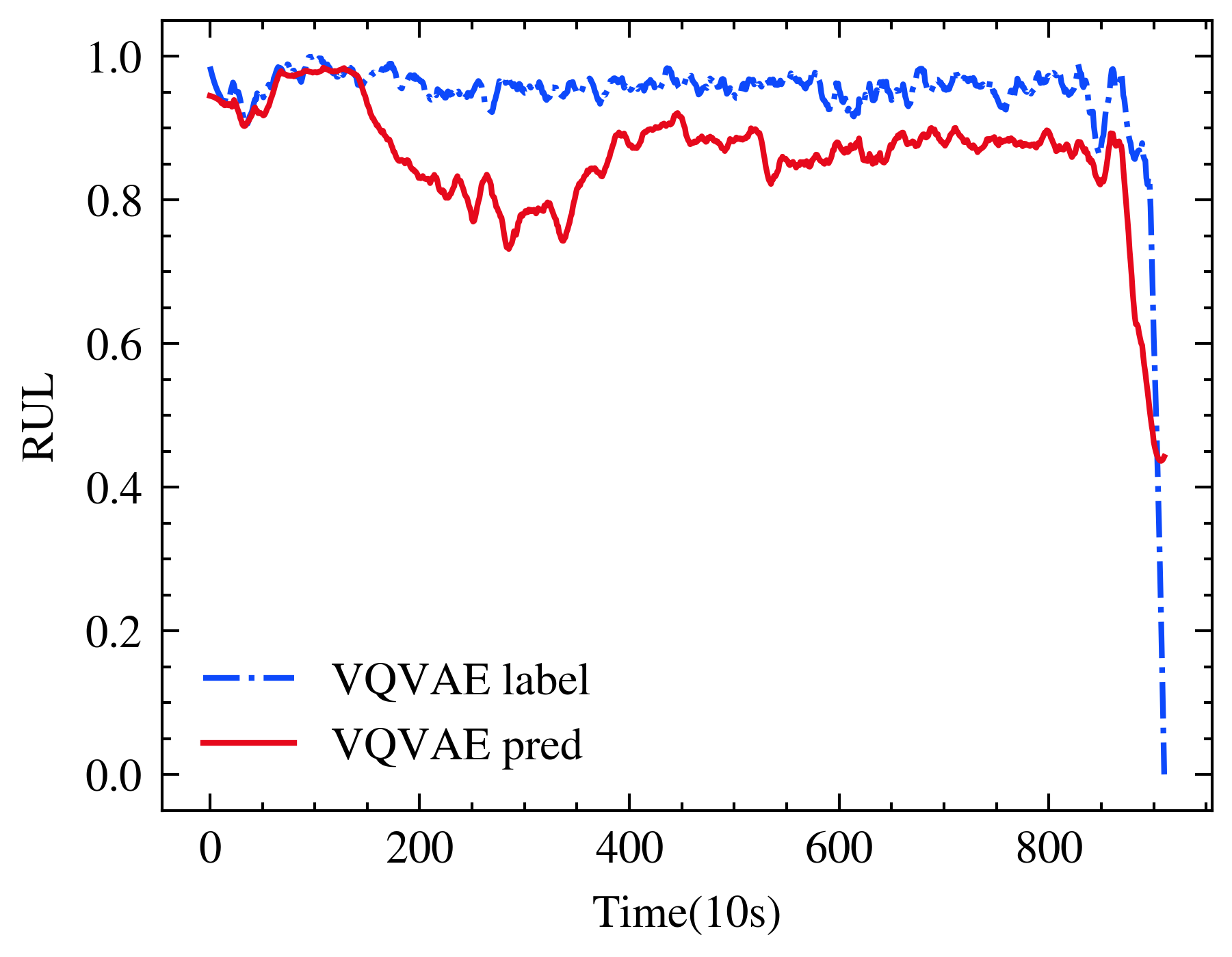} 
        \\
        (a)  & (b) & (c) \\
    \end{tabularx}
    \caption{Prediction results of bearing 2-1. (a) Horizontal vibration of the original waveform. (b) vertical vibration of the original waveform. (c) Predicted and actual results of VQ-VAE.}
    \label{fig:pred2-1}
    \end{adjustwidth}
    
\end{figure}

\begin{figure}[H]
    \begin{adjustwidth}{-\extralength}{0cm}
   \begin{tabularx}{\linewidth}{CCC}
        \includegraphics[width=\linewidth]{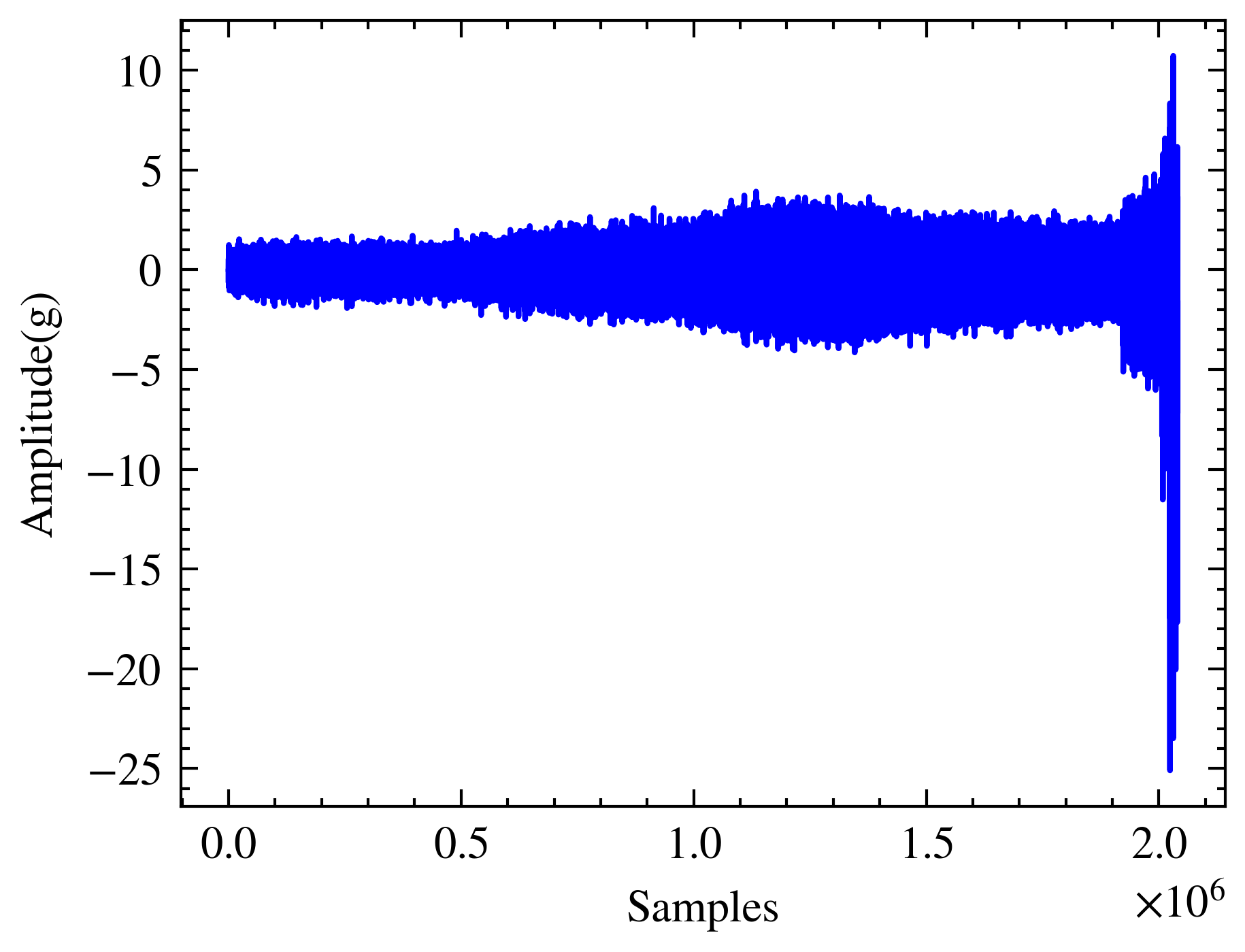} & \includegraphics[width=\linewidth]{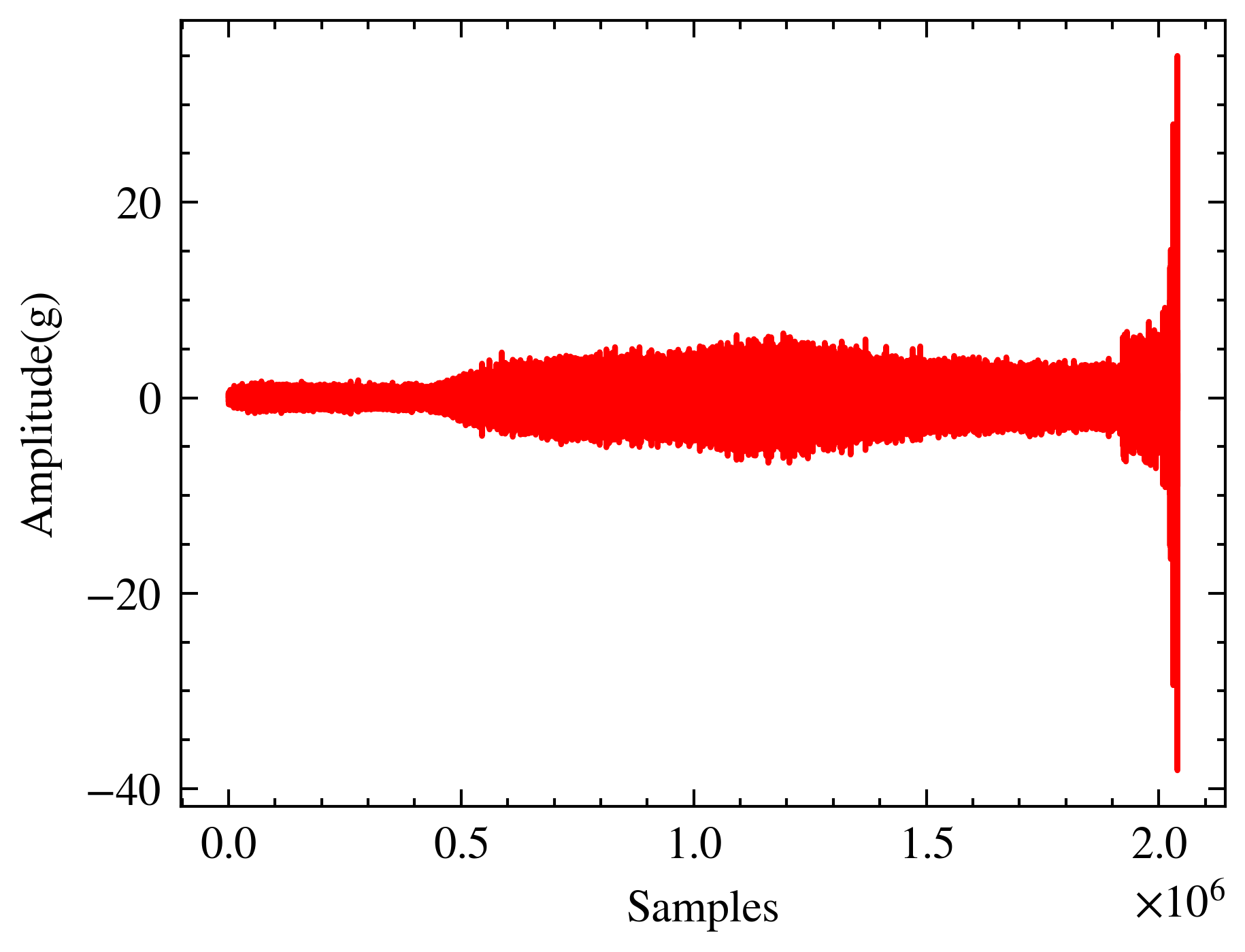} &
        \includegraphics[width=\linewidth]{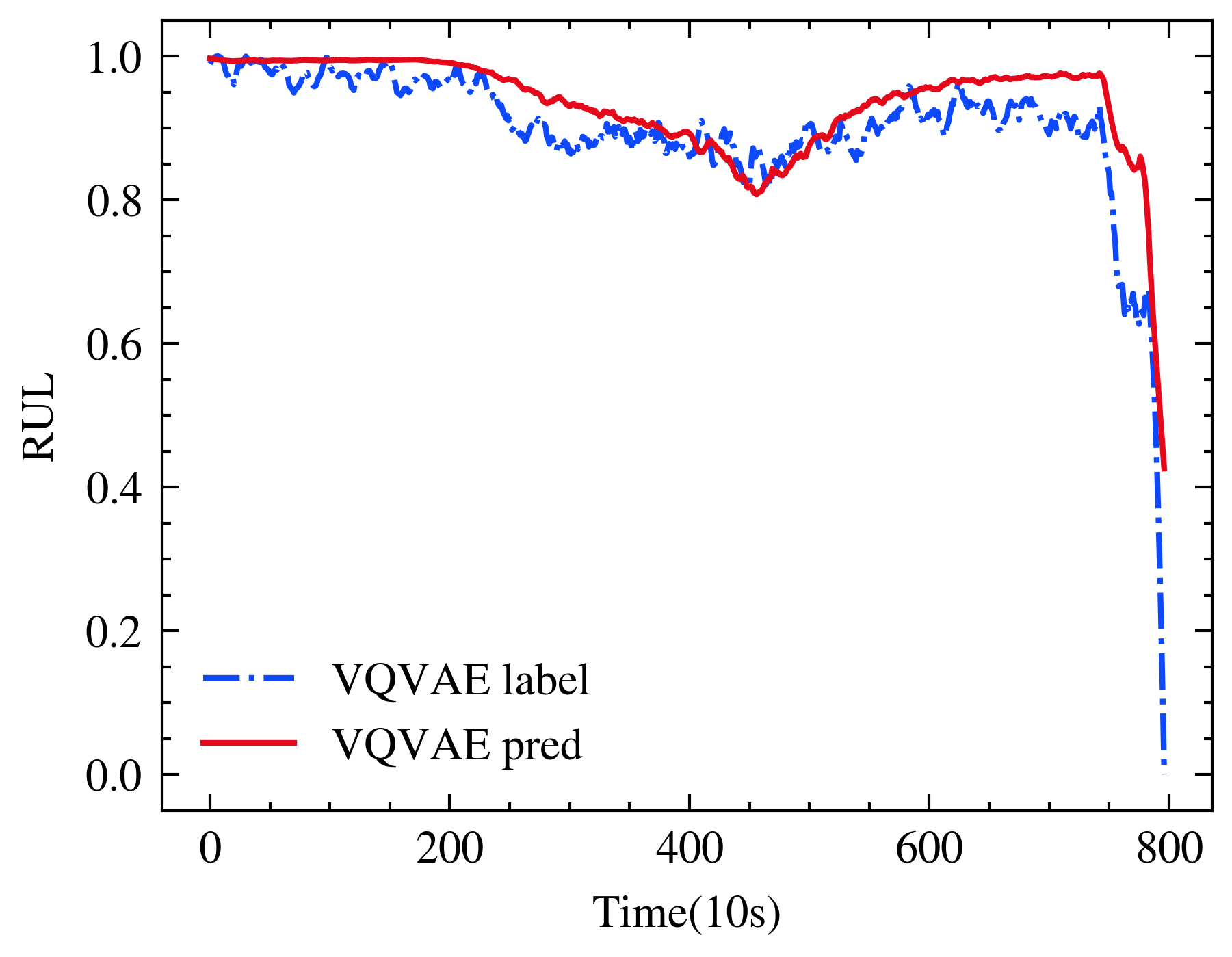} 
        \\
        (a)  & (b) & (c) \\
    \end{tabularx}
    \caption{Prediction results of bearing 2-2. (a) Horizontal vibration of the original waveform. (b) vertical vibration of the original waveform. (c) Predicted and actual results of VQ-VAE.}
    \label{fig:pred2-2}
    \end{adjustwidth}
\end{figure}

\begin{figure}[H]
     \begin{adjustwidth}{-\extralength}{0cm}
      \begin{tabularx}{\linewidth}{CCC}
        \includegraphics[width=\linewidth]{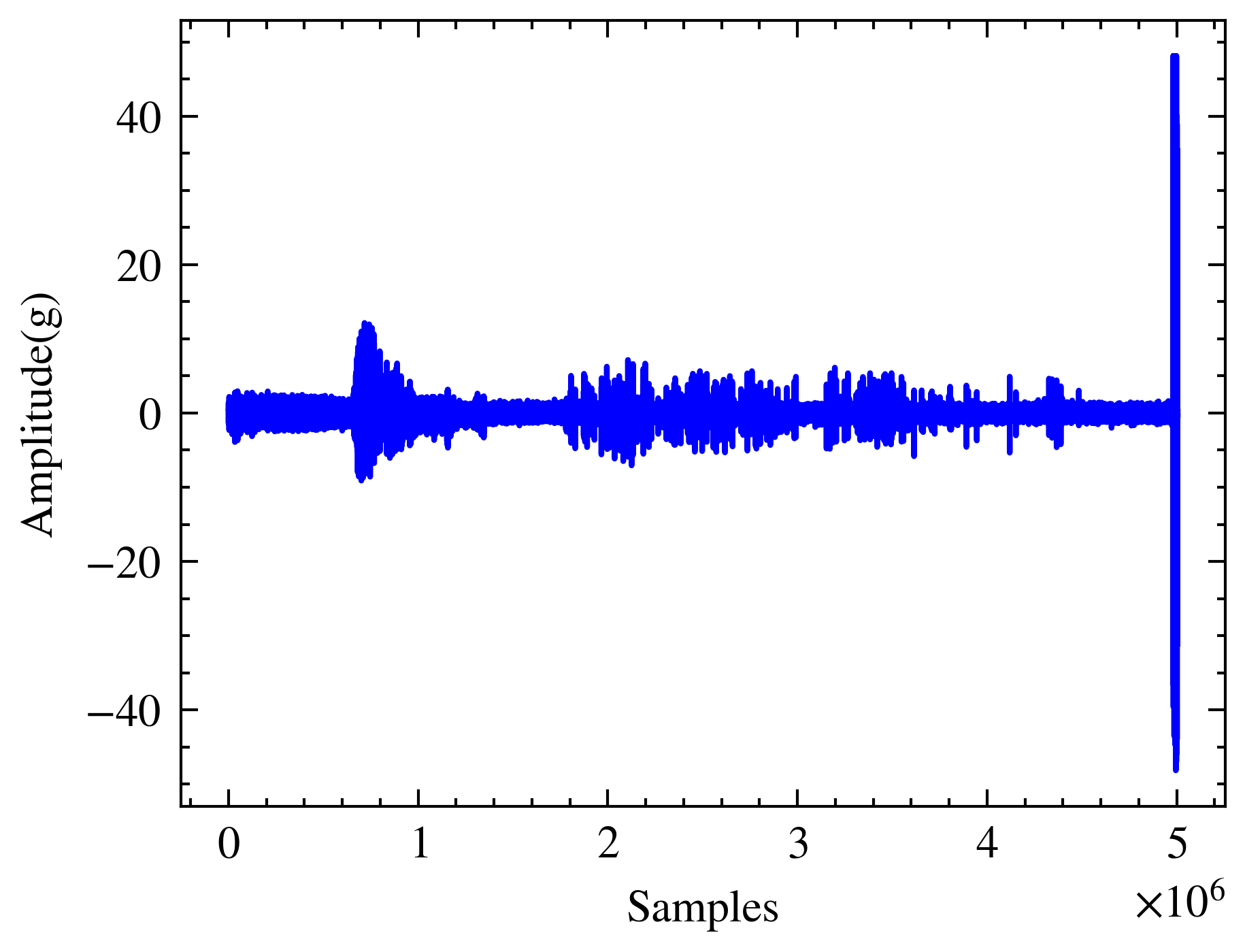} & \includegraphics[width=\linewidth]{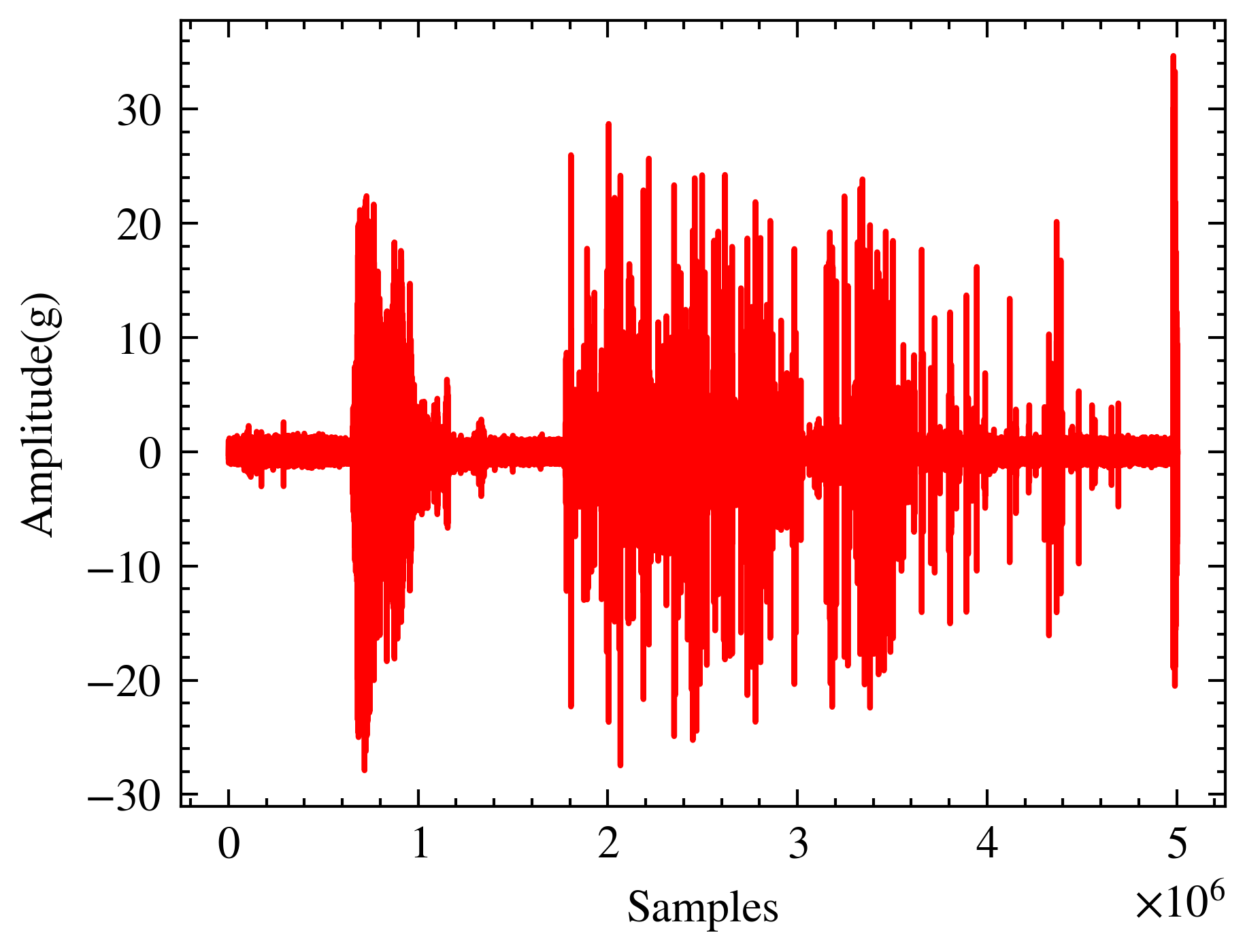} &
        \includegraphics[width=\linewidth]{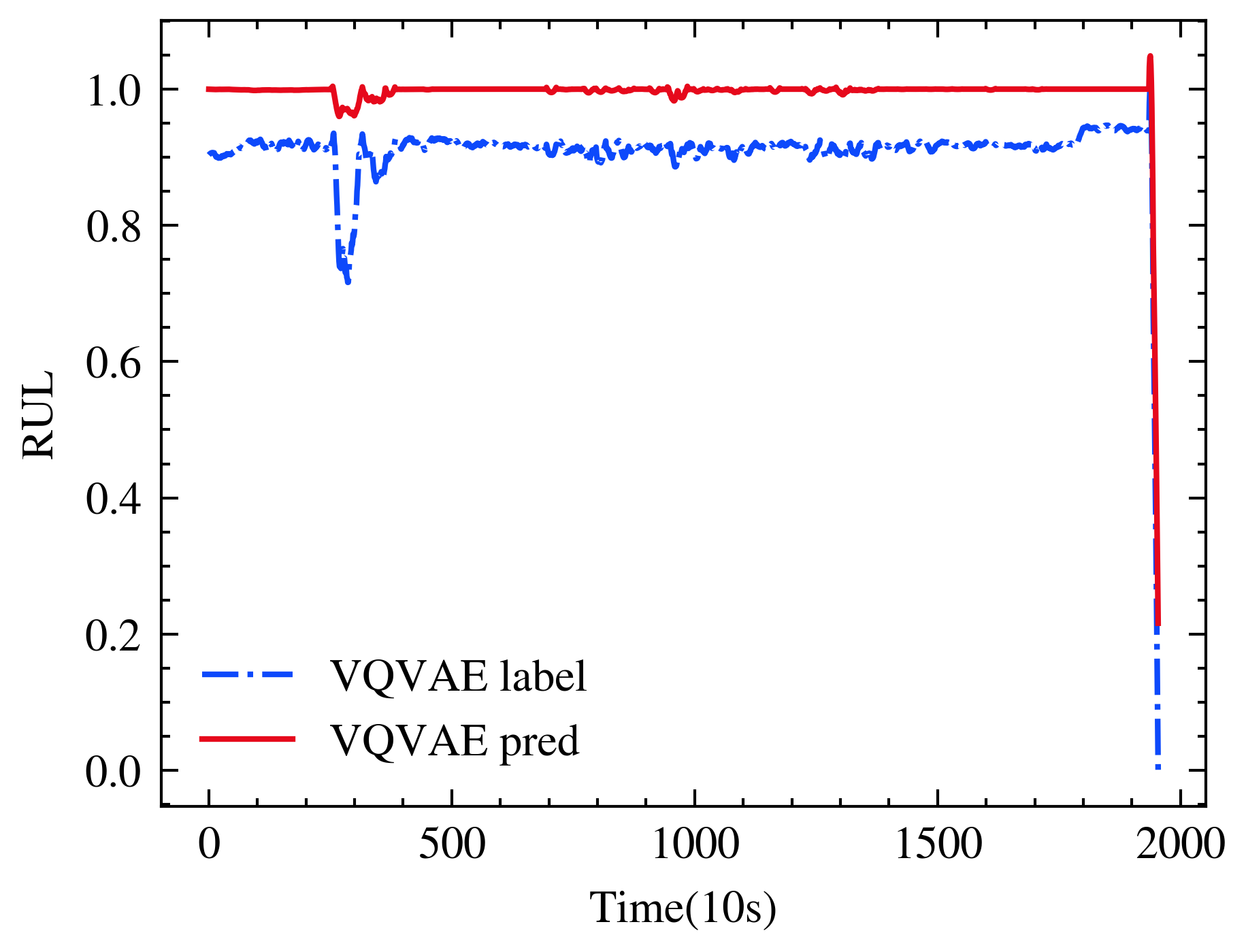} 
        \\
        (a)  & (b) & (c) \\
    \end{tabularx}
    \caption{Prediction results of bearing 2-3. (a) Horizontal vibration of the original waveform. (b) vertical vibration of the original waveform. (c) Predicted and actual results of VQ-VAE.}
    \label{fig:pred2-3}
    \end{adjustwidth}
\end{figure}

\begin{figure}[H]
    \begin{adjustwidth}{-\extralength}{0cm}
      \begin{tabularx}{\linewidth}{CCC}
        \includegraphics[width=\linewidth]{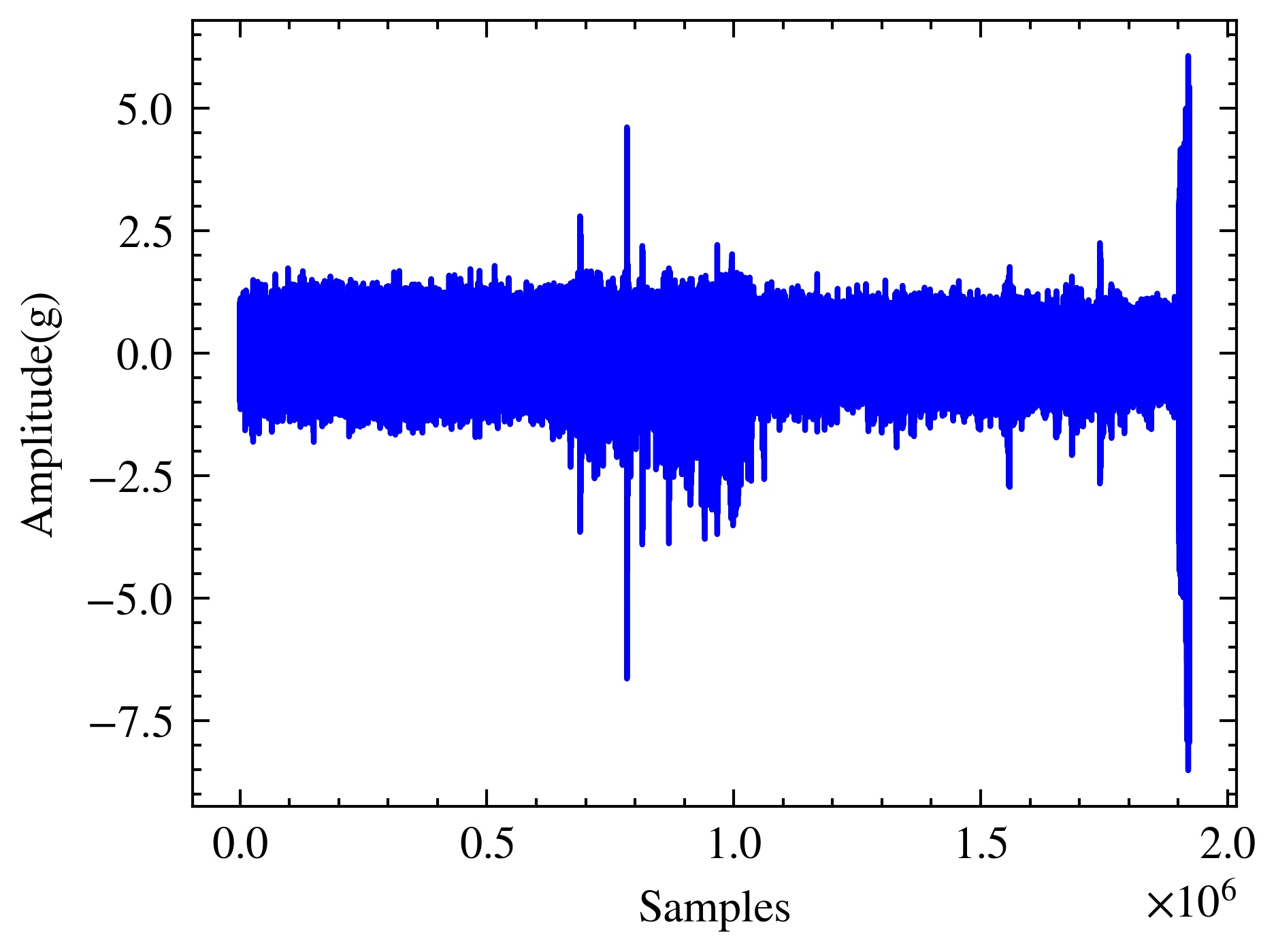} & \includegraphics[width=\linewidth]{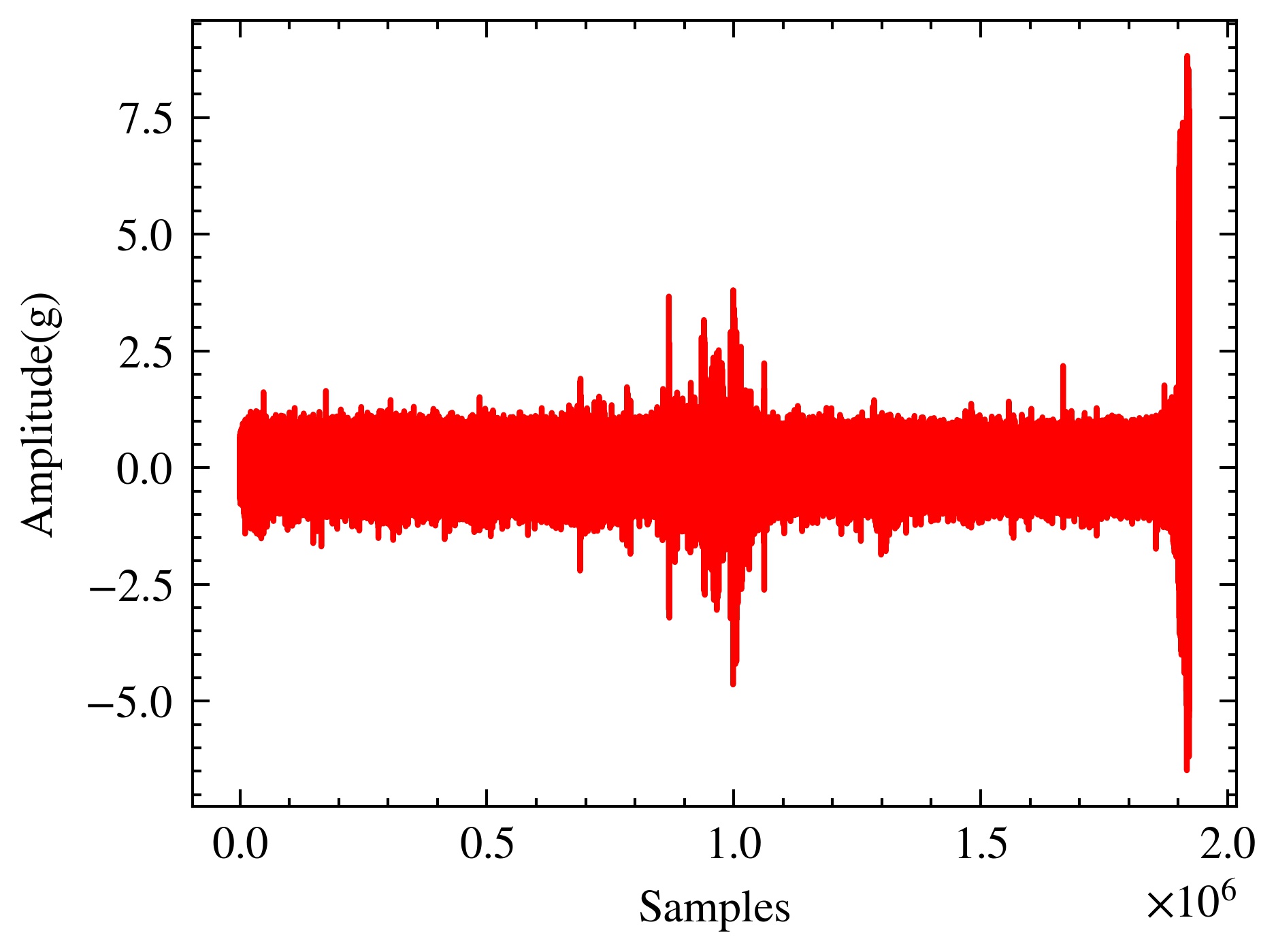} &
        \includegraphics[width=\linewidth]{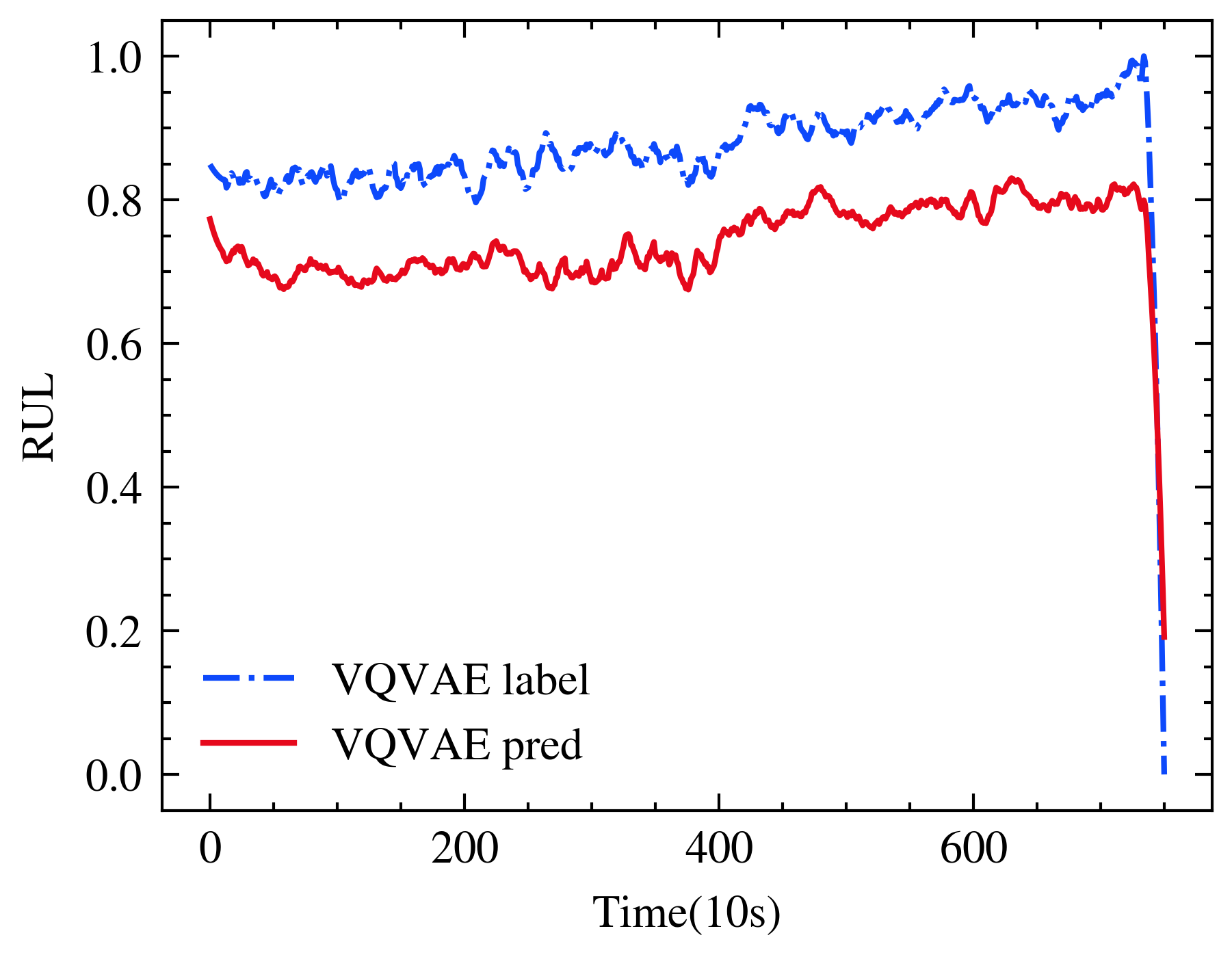} 
        \\
        (a)  & (b) & (c) \\
    \end{tabularx}
    \caption{Prediction results of bearing 2-4. (a) Horizontal vibration of the original waveform. (b) vertical vibration of the original waveform. (c) Predicted and actual results of VQ-VAE.}
    \label{fig:pred2-4}
    \end{adjustwidth}
\end{figure}

A proficient model often exhibits commendable performance in cross-domain scenarios. Therefore, under condition 2, bearings 2-1 to 2-4 are selected as the test set. The AE, VQ-VAE, and ASTCN models, having been trained on bearings 1-1 to 1-6, are directly tested without fine-tuning. Both prediction models and label models show good generalization ability, with the results illustrated in Figures \ref{fig:pred2-1}-\ref{fig:pred2-4}, which displays the performance metrics across different test bearings.".

In Figure \ref{fig:pred2-1}, a trough is observed in the prediction results of ASTCN within the interval of t=250, while the prediction values of VQ-VAE exhibits a more stable performance. Similar phenomena are encountered in Figures \ref{fig:pred2-2} and \ref{fig:pred2-3}, where both bearing 2-2 and 2-3 display a concurrent trough in labels and prediction results. This arises as both the label and prediction models lack the capability for temporal modeling. In Figure \ref{fig:pred2-4}, a certain deviation in initial values between the label model and prediction model is noted, along with issues concerning trend direction. This problem could be ameliorated through cross-domain methodologies.

\section{Conclusions}\label{section:con}
This paper proposes an end-to-end health indicator (HI) construction method using vector quantised variational autoencoder (VQ-VAE) for predicting the remaining useful life (RUL) of bearings. This method addresses the issue of suboptimal HI metrics generated by traditional statistical methods such as RMS and kurtosis and resolves the need for dimensionality reduction of latent variables in traditional unsupervised learning methods such as Autoencoder. Through empirical testing, it is demonstrated that traditional statistical metrics do not adequately reflect the fluctuation in the curves. Consequently, two novel statistical metrics, mean absolute distance (MAD) and mean variance (MV), are introduced. These metrics accurately depict the fluctuation patterns in the curves, thereby reflecting the model's accuracy in judging similar features. Both label and prediction models were tested using these metrics. On the PMH2012 dataset, both the direct application of VQ-VAE and the combined method of feature selection followed by VQ-VAE for label construction achieved lower MAD and MV values. Furthermore, the ASTCN prediction model trained with VQ-VAE labels also exhibited promising results, attaining the lowest values for MAD and MV.

In subsequent work, the network architecture of VQ-VAE will continue to be optimized to reduce the up-and-down fluctuations in the curves, with the aim of constructing a more accurate label for training. Consideration will be given to employing some cross-domain methods to enhance its generalization ability under different operating conditions or even across different datasets.

\authorcontributions{Conceptualization, J.W. and Q.Z.; methodology, J.W. and Q.Z.; software, G.Z.; validation, J.W. and G.Z.; formal analysis, G.S.; investigation, J.W. and Q.Z.; resources, G.S.; data curation, G.Z.; writing---original draft preparation, J.W.; writing---review and editing, G.Z.; visualization, G.Z.; supervision, Q.Z.; project administration, Q.Z. and G.S.; funding acquisition, G.S. All authors have read and agreed to the published version of the manuscript.}

\funding{The research was partially supported by the Key Project of Natural Science Foundation of China (No. 61933013), the Science and Technology Innovation Strategy Project of Guangdong Provincial(No. 2023S002028) and the special projects in key fields of ordinary universities in Guangdong Province(No.2023ZDZX3015)}

\informedconsent{Not applicable}

\dataavailability{The datasets used in this paper are publicly available to everyone and
can be accessed at: https://github.com/Lucky-Loek/ieee-phm-2012-data-challenge-dataset} 

\acknowledgments{The research was partially supported by the Key Project of Natural Science Foundation of China (No. 61933013), the Science and Technology Innovation Strategy Project of Guangdong Provincial(No. 2023S002028) and the special projects in key fields of ordinary universities in Guangdong Province(No.2023ZDZX3015). We also appreciate the technical support of the Guangdong Provincial Key Laboratory of Petrochemical Equipment and Fault Diagnosis.  }

\conflictsofinterest{The authors declare no conflict of interest.} 

\appendixtitles{no} 
\appendixstart
\appendix
\section[\appendixname~\thesection]{}


\begin{table}[htbp] \centering \caption{The Encoder structure of the Autoencoder. } \label{tab:autoencoder} \begin{tabularx}{\textwidth}{CCCCC} \toprule Layer name & Kernels size & Stride & Kernels number & Output size \\ \midrule Input & – & – & – & 1024 x 1 \\ Convolution1 & 1 x 3 & 1 x 1 & 16 & 1024 x 16 \\ Pooling1 & 1 x 2 & 1 x 2 & – & 512 x 16 \\ Convolution2 & 1 x 3 & 1 x 1 & 32 & 512 x 32 \\ Pooling2 & 1 x 2 & 1 x 2 & – & 256 x 32 \\ Convolution3 & 1 x 3 & 1 x 1 & 64 & 256 x 64 \\ Pooling3 & 1 x 4 & 1 x 4 & – & 64 x 64 \\ Convolution4 & 1 x 3 & 1 x 1 & 128 & 64 x 128 \\ Pooling4 & 1 x 2 & 1 x 2 & – & 32 x 128 \\ Convolution5 & 1 x 3 & 1 x 1 & 32 & 32 x 32 \\ Convolution6 & 1 x 3 & 1 x 1 & 1 & 32 x 1  \\ \bottomrule \end{tabularx}  \end{table}

\begin{table}[htbp] \centering \caption{The Encoder structure of the "feature+AE". } \label{tab:small-autoencoder} \begin{tabularx}{\textwidth}{CCCCC} \toprule Layer name & Kernels size & Stride & Kernels number & Output size \\ \midrule Input & – & – & – & 38 x 1 \\ Convolution1 & 1 x 3 & 1 x 1 & 16 & 36 x 16  \\ Pooling1 & 1 x 2 & 1 x 2 & – & 18 x 16 \\Convolution2 & 1 x 3 & 1 x 1 & 32 & 18 x 32 \\   Pooling2 & 1 x 2 & 1 x 2 & – & 9 x 32\\ Convolution3 & 1 x 3 & 1 x 1 & 64 & 9 x 64  \\ Convolution4 & 1 x 3 & 1 x 1 & 128 & 9 x 128 \\   Convolution5 & 1 x 3 & 1 x 1 & 32 & 9 x 32 \\ Convolution6 & 1 x 3 & 1 x 1 & 1 & 9 x 1  \\ \bottomrule \end{tabularx}  \end{table}

\begin{table}[htbp] \centering \caption{The Encoder structure of the "feature+VQVAE". } \label{tab:small-VQVAE} \begin{tabularx}{\textwidth}{CCCCC} \toprule Layer name & Kernels size & Stride & Kernels number & Output size \\ \midrule Input & – & – & – & 38 x 1 \\ Convolution1 & 1 x 3 & 1 x 1 & 4 & 36 x 4  \\ Pooling1 & 1 x 2 & 1 x 2 & – & 18 x 4 \\Convolution2 & 1 x 3 & 1 x 1 & 4 & 18 x 4 \\   Pooling2 & 1 x 2 & 1 x 2 & – & 9 x 4\\ Convolution3 & 1 x 3 & 1 x 1 & 8 & 9 x 8  \\ Convolution4 & 1 x 3 & 1 x 1 & 16 & 9 x 16 \\   Convolution5 & 1 x 3 & 1 x 1 & 4 & 9 x 4 \\  \bottomrule \end{tabularx}  \end{table}

\begin{table}[H]
    \begin{adjustwidth}{-\extralength}{0cm}
    \caption{Comparative analysis of the definite results of label models for bearings 1 through 7. \label{tab:labels-all}}
     \begin{tabularx}{\linewidth}{XCCCCCCCCC}
    \toprule
        \textbf{Method} & \textbf{} & \textbf{Bearing1-1} & \textbf{Bearing1-2} & \textbf{Bearing1-3} & \textbf{Bearing1-4} & \textbf{Bearing1-5} & \textbf{Bearing1-6} & \textbf{Bearing1-7}  \\ \midrule
        AE & Mon & 0.0143  & 0.0437  & 0.0194  & 0.0708  & 0.0219  & 0.0176  & 0.0292    \\ 
        ~ & Tend & -0.376  & -0.245  & -0.552  & -0.668  & -0.009  & -0.049  & -0.268    \\ 
        ~ & MAD & 0.0174  & 0.0429  & 0.0070  & 0.0353  & 0.0314  & 0.0387  & 0.0322    \\ 
        ~ & MV & 0.00034  & 0.00190  & 0.00041  & 0.00147  & 0.00080  & 0.00136  & 0.00096    \\ \midrule
        Feature+VQVAE & Mon & 0.0128  & 0.0000  & 0.0093  & 0.0329  & 0.0268  & 0.0298  & 0.0301    \\ 
        ~ & Tend & -0.268  & -0.231  & -0.463  & -0.523  & -0.339  & -0.123  & -0.095    \\ 
        ~ & MAD & 0.0051  & 0.0326  & 0.0190  & 0.0196  & 0.0199  & 0.0256  & 0.0290    \\ 
        ~ & MV & 0.00022  & 0.00282  & 0.00068  & 0.00065  & 0.00058  & 0.00356  & 0.00095    \\ \midrule
        VQVAE(Our) & Mon & 0.0371  & 0.0621  & 0.0312  & 0.0399  & 0.0041  & 0.0086  & 0.0106   \\ 
        ~ & Tend & -0.525  & -0.393  & -0.558  & -0.729  & 0.212  & 0.002  & -0.231    \\ 
        ~ & MAD & 0.0202  & 0.0454  & 0.0222  & 0.0160  & 0.0406  & 0.0474  & 0.0253    \\ 
        ~ & MV & 0.00045  & 0.00204  & 0.00085  & 0.00050  & 0.00129  & 0.00183  & 0.00060    \\  \midrule
        Feature+AE & Mon & 0.0121  & 0.0276  & 0.0447  & 0.0007  & 0.0292  & 0.0061  & 0.0027    \\ 
        ~ & Tend & -0.596  & -0.033  & -0.691  & -0.617  & -0.188  & -0.031  & -0.156    \\ 
        ~ & MAD & 0.0159  & 0.0730  & 0.0402  & 0.0352  & 0.0257  & 0.0418  & 0.0289    \\ 
        ~ & MV & 0.00054  & 0.00612  & 0.00163  & 0.00093  & 0.00079  & 0.00354  & 0.00081    \\ \midrule
        RMS & Mon & 0.0264  & 0.0713  & 0.0455  & 0.0834  & 0.0471  & 0.0037  & 0.0292    \\ 
        ~ & Tend & -0.383  & -0.145  & -0.421  & -0.714  & -0.185  & -0.270  & -0.741    \\ 
        ~ & MAD & 0.0084  & 0.1423  & 0.0160  & 0.0100  & 0.0143  & 0.0294  & 0.0146    \\ 
        ~ & MV & 0.00020  & 0.02114  & 0.00068  & 0.00023  & 0.00056  & 0.00408  & 0.00025    \\  \midrule
        PCA & Mon & 0.0264  & 0.0575  & 0.0101  & 0.0638  & 0.0203  & 0.0135  & 0.0735    \\ 
        ~ & Tend & -0.549  & -0.318  & -0.849  & -0.693  & 0.077  & -0.247  & 0.848   \\ 
        ~ & MAD & 0.0132  & 0.1869  & 0.0665  & 0.0068  & 0.0103  & 0.0220  & 0.0295    \\ 
        ~ & MV & 0.00026  & 0.03346  & 0.00677  & 0.00009  & 0.00028  & 0.00166  & 0.00061   \\ \bottomrule
    \end{tabularx}
    
     \end{adjustwidth}
\end{table}

\begin{table}[H]
     \begin{adjustwidth}{-\extralength}{0cm}
      \caption{Comparative analysis of the definite results of prediction models for bearings 1 through 7. \label{tab:preds-all}}
    \begin{tabularx}{\linewidth}{XCCCCCCCCC}
    \toprule
        \textbf{Method} & ~ & \textbf{Bearing1-1} & \textbf{Bearing1-2} & \textbf{Bearing1-3} & \textbf{Bearing1-4} & \textbf{Bearing1-5} & \textbf{Bearing1-6} & \textbf{Bearing1-7} \\ \midrule
        Featuer+AE & RMSE & 0.060  & 0.144  & 0.174  & 0.077  & 0.178  & 0.074  & 0.191  \\ 
        ~ & MAE & 0.040  & 0.117  & 0.084  & 0.061  & 0.144  & 0.061  & 0.168  \\ 
        ~ & MAD & 0.0023  & 0.0525  & 0.0040  & 0.0234  & 0.1567  & 0.0358  & 0.0530  \\ 
        ~ & MV & 0.00043  & 0.00199  & 0.00008  & 0.00327  & 0.02137  & 0.00307  & 0.00368  \\ 
        ~ & Score & 9.1  & 22.5  & 9.5  & 5.3  & 27.3  & 16.6  & 35.2  \\ \midrule
        AE & RMSE & 0.069  & 0.127  & 0.230  & 0.064  & 0.111  & 0.262  & 0.229  \\ 
        ~ & MAE & 0.042  & 0.118  & 0.220  & 0.038  & 0.097  & 0.256  & 0.221  \\ 
        ~ & MAD & 0.0029  & 0.0570  & 0.0495  & 0.1004  & 0.1003  & 0.0647  & 0.0278  \\ 
        ~ & MV & 0.00032  & 0.00323  & 0.00192  & 0.01059  & 0.00876  & 0.00441  & 0.00134  \\ 
        ~ & Score & 9.5  & 22.7  & 24.5  & 3.0  & 18.7  & 56.0  & 53.0  \\ \midrule
        VQVAE(Our) & RMSE & 0.084  & 0.094  & 0.166  & 0.067  & 0.094  & 0.055  & 0.046  \\ 
        ~ & MAE & 0.077  & 0.085  & 0.086  & 0.053  & 0.082  & 0.037  & 0.020  \\ 
        ~ & MAD & 0.0442  & 0.0443  & 0.0044  & 0.0294  & 0.0423  & 0.0135  & 0.0145  \\ 
        ~ & MV & 0.00141  & 0.00174  & 0.00014  & 0.00185  & 0.00185  & 0.00075  & 0.00082  \\ 
        ~ & Score & 13.6  & 16.3  & 10.6  & 4.4  & 20.3  & 10.5  & 4.3  \\  \midrule
        Feature+VQVAE & RMSE & 0.055  & 0.066  & 0.124  & 0.079  & 0.106  & 0.037  & 0.071  \\ 
        ~ & MAE & 0.039  & 0.052  & 0.100  & 0.064  & 0.093  & 0.023  & 0.051  \\ 
        ~ & MAD & 0.0525  & 0.0537  & 0.0145  & 0.0770  & 0.0137  & 0.0045  & 0.0481  \\ 
        ~ & MV & 0.00187  & 0.00188  & 0.00066  & 0.01020  & 0.00190  & 0.00029  & 0.00757  \\ 
        ~ & Score & 8.4  & 10.0  & 13.9  & 5.5  & 22.9  & 6.3  & 12.1  \\  \midrule
        Linear & RMSE & 0.280  & 0.269  & 0.140  & 0.283  & 0.254  & 0.130  & 0.178  \\ 
        ~ & MAE & 0.225  & 0.231  & 0.106  & 0.214  & 0.188  & 0.097  & 0.143  \\ 
        ~ & MAD & 0.0715  & 0.0649  & 0.0759  & 0.1331  & 0.1855  & 0.0780  & 0.0926  \\ 
        ~ & MV & 0.00385  & 0.00288  & 0.00468  & 0.01679  & 0.02267  & 0.00514  & 0.00586  \\ 
        ~ & Score & 48.4  & 47.0  & 14.1  & 18.0  & 43.9  & 23.0  & 29.5  \\  \midrule
        Piecewise & RMSE & 0.070  & 0.060  & 0.226  & 0.129  & 0.765  & 0.128  & 0.095  \\ 
        ~ & MAE & 0.010  & 0.035  & 0.100  & 0.027  & 0.713  & 0.086  & 0.041  \\ 
        ~ & MAD & 0.0022  & 0.0267  & 0.0015  & 0.0021  & 0.0485  & 0.0782  & 0.0420  \\ 
        ~ & MV & 0.00045  & 0.00127  & 0.00031  & 0.00076  & 0.00173  & 0.00800  & 0.00400  \\ 
        ~ & Score & 2.2  & 6.8  & 11.3  & 2.4  & 138.9  & 19.6  & 9.3  \\ \midrule
        PCA & RMSE & 0.162  & 0.420  & 0.110  & 0.327  & 0.100  & 0.088  & 0.288  \\ 
        ~ & MAE & 0.134  & 0.417  & 0.052  & 0.292  & 0.058  & 0.056  & 0.194  \\ 
        ~ & MAD & 0.0864  & 0.1140  & 0.0111  & 0.0735  & 0.0344  & 0.0280  & 0.0487  \\ 
        ~ & MV & 0.00512  & 0.00934  & 0.00065  & 0.00635  & 0.01234  & 0.00551  & 0.00570  \\ 
        ~ & Score & 23.7  & 80.4  & 6.6  & 25.9  & 13.4  & 15.0  & 47.1 \\ \bottomrule
    \end{tabularx}
   
     \end{adjustwidth}
\end{table}

\begin{adjustwidth}{-\extralength}{0cm}

\reftitle{References}


\bibliography{ciation}

\begin{thebibliography}{999}

\bibitem[Alzhrani and Atallah(2022)]{ref1}
Alzhrani, A.; Atallah, K.
\newblock Novel Passive Electrodynamic Magnetic Bearings.
\newblock In Proceedings of the 2022 IEEE Energy Conversion Congress and Exposition (ECCE),  oct 2022, pp. 1--7.
\newblock {\url{https://doi.org/10.1109/ECCE50734.2022.9947987}}.

\bibitem[Prasad and Narayanan(2019)]{ref2}
Prasad, K.N.V.; Narayanan, G.
\newblock Electro-Magnetic Bearings with Power Electronic Control for High-Speed Rotating Machines: A Review.
\newblock In Proceedings of the 2019 National Power Electronics Conference (NPEC),  dec 2019, pp. 1--6.
\newblock {\url{https://doi.org/10.1109/NPEC47332.2019.9034789}}.

\bibitem[Valeev et~al.(2020)Valeev, Tokarev, and Zotov]{ref3}
Valeev, A.; Tokarev, A.; Zotov, A.
\newblock Diagnostics of Bearings of Industrial Machines Using Real-Time Strain Gauge Analysis.
\newblock In Proceedings of the 2020 International Conference on Industrial Engineering, Applications and Manufacturing (ICIEAM),  may 2020, pp. 1--5.
\newblock {\url{https://doi.org/10.1109/ICIEAM48468.2020.9111963}}.

\bibitem[An et~al.(2015)An, Kim, and Choi]{ref4}
An, D.; Kim, N.H.; Choi, J.H.
\newblock Practical Options for Selecting Data-Driven or Physics-Based Prognostics Algorithms with Reviews.
\newblock {\em Reliability Engineering \& System Safety} {\bf 2015}, {\em 133},~223--236.
\newblock {\url{https://doi.org/10.1016/j.ress.2014.09.014}}.

\bibitem[Singleton et~al.(2017)Singleton, Strangas, and Aviyente]{ref5}
Singleton, R.K.; Strangas, E.G.; Aviyente, S.
\newblock The Use of Bearing Currents and Vibrations in Lifetime Estimation of Bearings.
\newblock {\em IEEE Transactions on Industrial Informatics} {\bf 2017}, {\em 13},~1301--1309.
\newblock {\url{https://doi.org/10.1109/TII.2016.2643693}}.

\bibitem[Qian and Yan(2015)]{ref6}
Qian, Y.; Yan, R.
\newblock Remaining Useful Life Prediction of Rolling Bearings Using an Enhanced Particle Filter.
\newblock {\em IEEE Transactions on Instrumentation and Measurement} {\bf 2015}, {\em 64},~2696--2707.
\newblock {\url{https://doi.org/10.1109/TIM.2015.2427891}}.

\bibitem[Miao et~al.(2019)Miao, Li, Sun, and Liu]{ref7}
Miao, H.; Li, B.; Sun, C.; Liu, J.
\newblock Joint Learning of Degradation Assessment and RUL Prediction for Aeroengines via Dual-Task Deep LSTM Networks.
\newblock {\em IEEE Transactions on Industrial Informatics} {\bf 2019}, {\em 15},~5023--5032.
\newblock {\url{https://doi.org/10.1109/TII.2019.2900295}}.

\bibitem[Chen et~al.(2019)Chen, Gryllias, and Li]{ref8}
Chen, Z.; Gryllias, K.; Li, W.
\newblock Mechanical Fault Diagnosis Using Convolutional Neural Networks and Extreme Learning Machine.
\newblock {\em Mechanical Systems and Signal Processing} {\bf 2019}, {\em 133},~106272.
\newblock {\url{https://doi.org/10.1016/j.ymssp.2019.106272}}.

\bibitem[Ren et~al.(2018)Ren, Sun, Wang, and Zhang]{ref16}
Ren, L.; Sun, Y.; Wang, H.; Zhang, L.
\newblock Prediction of Bearing Remaining Useful Life With Deep Convolution Neural Network.
\newblock {\em IEEE Access} {\bf 2018}, {\em 6},~13041--13049.
\newblock {\url{https://doi.org/10.1109/ACCESS.2018.2804930}}.

\bibitem[Yuan et~al.(2016)Yuan, Wu, and Lin]{ref17}
Yuan, M.; Wu, Y.; Lin, L.
\newblock Fault Diagnosis and Remaining Useful Life Estimation of Aero Engine Using LSTM Neural Network.
\newblock In Proceedings of the 2016 IEEE International Conference on Aircraft Utility Systems (AUS),  oct 2016, pp. 135--140.
\newblock {\url{https://doi.org/10.1109/AUS.2016.7748035}}.

\bibitem[Que et~al.(2021)Que, Jin, and Xu]{ref18}
Que, Z.; Jin, X.; Xu, Z.
\newblock Remaining Useful Life Prediction for Bearings Based on a Gated Recurrent Unit.
\newblock {\em IEEE Transactions on Instrumentation and Measurement} {\bf 2021}, {\em 70},~1--11.
\newblock {\url{https://doi.org/10.1109/TIM.2021.3054025}}.

\bibitem[Bai et~al.(2018)Bai, Kolter, and Koltun]{ref19}
Bai, S.; Kolter, J.Z.; Koltun, V.
\newblock An Empirical Evaluation of Generic Convolutional and Recurrent Networks for Sequence Modeling,  2018,  \href{http://xxx.lanl.gov/abs/1803.01271}{{\normalfont [1803.01271]}}.
\newblock {\url{https://doi.org/10.48550/arXiv.1803.01271}}.

\bibitem[Sun et~al.(2021)Sun, Xia, Hu, Lu, Liu, and Wang]{ref20}
Sun, H.; Xia, M.; Hu, Y.; Lu, S.; Liu, Y.; Wang, Q.
\newblock A New Sorting Feature-Based Temporal Convolutional Network for Remaining Useful Life Prediction of Rotating Machinery.
\newblock {\em Computers and Electrical Engineering} {\bf 2021}, {\em 95},~107413.
\newblock {\url{https://doi.org/10.1016/j.compeleceng.2021.107413}}.

\bibitem[Gan et~al.(2021)Gan, Li, Zhou, and Tang]{ref21}
Gan, Z.; Li, C.; Zhou, J.; Tang, G.
\newblock Temporal Convolutional Networks Interval Prediction Model for Wind Speed Forecasting.
\newblock {\em Electric Power Systems Research} {\bf 2021}, {\em 191},~106865.
\newblock {\url{https://doi.org/10.1016/j.epsr.2020.106865}}.

\bibitem[Shen et~al.(2012)Shen, He, Chen, Sun, and Liu]{ref9}
Shen, Z.; He, Z.; Chen, X.; Sun, C.; Liu, Z.
\newblock A Monotonic Degradation Assessment Index of Rolling Bearings Using Fuzzy Support Vector Data Description and Running Time.
\newblock {\em Sensors} {\bf 2012}, {\em 12},~10109--10135.
\newblock {\url{https://doi.org/10.3390/s120810109}}.

\bibitem[Soualhi et~al.(2014)Soualhi, Razik, Clerc, and Doan]{ref10}
Soualhi, A.; Razik, H.; Clerc, G.; Doan, D.D.
\newblock Prognosis of Bearing Failures Using Hidden Markov Models and the Adaptive Neuro-Fuzzy Inference System.
\newblock {\em IEEE Transactions on Industrial Electronics} {\bf 2014}, {\em 61},~2864--2874.
\newblock {\url{https://doi.org/10.1109/TIE.2013.2274415}}.

\bibitem[Gebraeel et~al.(2004)Gebraeel, Lawley, Liu, and Parmeshwaran]{ref11}
Gebraeel, N.; Lawley, M.; Liu, R.; Parmeshwaran, V.
\newblock Residual Life Predictions from Vibration-Based Degradation Signals: A Neural Network Approach.
\newblock {\em IEEE Transactions on Industrial Electronics} {\bf 2004}, {\em 51},~694--700.
\newblock {\url{https://doi.org/10.1109/TIE.2004.824875}}.

\bibitem[Benkedjouh et~al.(2013)Benkedjouh, Medjaher, Zerhouni, and Rechak]{ref12}
Benkedjouh, T.; Medjaher, K.; Zerhouni, N.; Rechak, S.
\newblock Remaining Useful Life Estimation Based on Nonlinear Feature Reduction and Support Vector Regression.
\newblock {\em Engineering Applications of Artificial Intelligence} {\bf 2013}, {\em 26},~1751--1760.
\newblock {\url{https://doi.org/10.1016/j.engappai.2013.02.006}}.

\bibitem[Dong and He(2007)]{ref13}
Dong, M.; He, D.
\newblock A Segmental Hidden Semi-Markov Model (HSMM)-Based Diagnostics and Prognostics Framework and Methodology.
\newblock {\em Mechanical Systems and Signal Processing} {\bf 2007}, {\em 21},~2248--2266.
\newblock {\url{https://doi.org/10.1016/j.ymssp.2006.10.001}}.

\bibitem[Djeziri et~al.(2018)Djeziri, Benmoussa, and Sanchez]{ref14}
Djeziri, M.A.; Benmoussa, S.; Sanchez, R.
\newblock Hybrid Method for Remaining Useful Life Prediction in Wind Turbine Systems.
\newblock {\em Renewable Energy} {\bf 2018}, {\em 116},~173--187.
\newblock {\url{https://doi.org/10.1016/j.renene.2017.05.020}}.

\bibitem[Cheng et~al.(2018)Cheng, Qu, and Qiao]{ref15}
Cheng, F.; Qu, L.; Qiao, W.
\newblock Fault Prognosis and Remaining Useful Life Prediction of Wind Turbine Gearboxes Using Current Signal Analysis.
\newblock {\em IEEE Transactions on Sustainable Energy} {\bf 2018}, {\em 9},~157--167.
\newblock {\url{https://doi.org/10.1109/TSTE.2017.2719626}}.

\bibitem[Guo et~al.(2017)Guo, Li, Jia, Lei, and Lin]{ref22}
Guo, L.; Li, N.; Jia, F.; Lei, Y.; Lin, J.
\newblock A Recurrent Neural Network Based Health Indicator for Remaining Useful Life Prediction of Bearings.
\newblock {\em Neurocomputing} {\bf 2017}, {\em 240},~98--109.
\newblock {\url{https://doi.org/10.1016/j.neucom.2017.02.045}}.

\bibitem[Kong and Yang(2019)]{ref23}
Kong, X.; Yang, J.
\newblock Remaining Useful Life Prediction of Rolling Bearings Based on RMS-MAVE and Dynamic Exponential Regression Model.
\newblock {\em IEEE Access} {\bf 2019}, {\em 7},~169705--169714.
\newblock {\url{https://doi.org/10.1109/ACCESS.2019.2954915}}.

\bibitem[Chen et~al.(2021)Chen, Qin, Wang, and Zhou]{ref24}
Chen, D.; Qin, Y.; Wang, Y.; Zhou, J.
\newblock Health Indicator Construction by Quadratic Function-Based Deep Convolutional Auto-Encoder and Its Application into Bearing RUL Prediction.
\newblock {\em ISA Transactions} {\bf 2021}, {\em 114},~44--56.
\newblock {\url{https://doi.org/10.1016/j.isatra.2020.12.052}}.

\bibitem[Liu et~al.(2021)Liu, Chen, Wen, Zhang, and Jiao]{ref25}
Liu, L.; Chen, J.; Wen, Z.; Zhang, D.; Jiao, L.
\newblock Densely Connected Fully Convolutional Auto-Encoder Based Slewing Bearing Degradation Trend Prediction Method.
\newblock In Proceedings of the 2021 Global Reliability and Prognostics and Health Management (PHM-Nanjing),  oct 2021, pp. 1--7.
\newblock {\url{https://doi.org/10.1109/PHM-Nanjing52125.2021.9612972}}.

\bibitem[{van den Oord} et~al.(2017){van den Oord}, Vinyals, and {kavukcuoglu}]{ref29}
{van den Oord}, A.; Vinyals, O.; {kavukcuoglu}, k.
\newblock Neural Discrete Representation Learning.
\newblock In Proceedings of the Advances in Neural Information Processing Systems. Curran Associates, Inc.,  2017, Vol.~30.

\bibitem[Chen et~al.(2020)Chen, Kang, Chen, and Wang]{ref26}
Chen, Y.; Kang, Y.; Chen, Y.; Wang, Z.
\newblock Probabilistic Forecasting with Temporal Convolutional Neural Network.
\newblock {\em Neurocomputing} {\bf 2020}, {\em 399},~491--501.
\newblock {\url{https://doi.org/10.1016/j.neucom.2020.03.011}}.

\bibitem[He et~al.(2016)He, Zhang, Ren, and Sun]{ref27}
He, K.; Zhang, X.; Ren, S.; Sun, J.
\newblock Deep Residual Learning for Image Recognition.
\newblock In Proceedings of the 2016 IEEE Conference on Computer Vision and Pattern Recognition (CVPR),  jun 2016, pp. 770--778.
\newblock {\url{https://doi.org/10.1109/CVPR.2016.90}}.

\bibitem[ref()]{ref28}
Autoencoder for Words - ScienceDirect.
\newblock https://www.sciencedirect.com/science/article/abs/pii/S0925231214003658.

\bibitem[Kingma and Welling(2022)]{ref30}
Kingma, D.P.; Welling, M.
\newblock Auto-Encoding Variational Bayes,  2022,  \href{http://xxx.lanl.gov/abs/1312.6114}{{\normalfont [1312.6114]}}.
\newblock {\url{https://doi.org/10.48550/arXiv.1312.6114}}.

\bibitem[Ince et~al.(2016)Ince, Kiranyaz, Eren, Askar, and Gabbouj]{ref40}
Ince, T.; Kiranyaz, S.; Eren, L.; Askar, M.; Gabbouj, M.
\newblock Real-Time Motor Fault Detection by 1-D Convolutional Neural Networks.
\newblock {\em IEEE Transactions on Industrial Electronics} {\bf 2016}, {\em 63},~7067--7075.
\newblock {\url{https://doi.org/10.1109/TIE.2016.2582729}}.

\bibitem[Wang et~al.(2022)Wang, Yang, Shi, and Wang]{ref31}
Wang, H.; Yang, J.; Shi, L.; Wang, R.
\newblock Remaining Useful Life Prediction Based on Adaptive SHRINKAGE Processing and Temporal Convolutional Network.
\newblock {\em Sensors} {\bf 2022}, {\em 22},~9088.
\newblock {\url{https://doi.org/10.3390/s22239088}}.

\bibitem[Wang et~al.(2021{\natexlab{a}})Wang, Jiang, Yang, and Zhang]{ref32}
Wang, C.; Jiang, W.; Yang, X.; Zhang, S.
\newblock RUL Prediction of Rolling Bearings Based on a DCAE and CNN.
\newblock {\em Applied Sciences} {\bf 2021}, {\em 11},~11516.
\newblock {\url{https://doi.org/10.3390/app112311516}}.

\bibitem[Wang et~al.(2021{\natexlab{b}})Wang, Shi, Hu, and Shen]{ref33}
Wang, R.; Shi, R.; Hu, X.; Shen, C.
\newblock Remaining Useful Life Prediction of Rolling Bearings Based on Multiscale Convolutional Neural Network with Integrated Dilated Convolution Blocks.
\newblock {\em Shock and Vibration} {\bf 2021}, {\em 2021},~e6616861.
\newblock {\url{https://doi.org/10.1155/2021/6616861}}.

\bibitem[Cao et~al.(2021)Cao, Jia, Ding, and Ding]{ref34}
Cao, Y.; Jia, M.; Ding, P.; Ding, Y.
\newblock Transfer Learning for Remaining Useful Life Prediction of Multi-Conditions Bearings Based on Bidirectional-GRU Network.
\newblock {\em Measurement} {\bf 2021}, {\em 178},~109287.
\newblock {\url{https://doi.org/10.1016/j.measurement.2021.109287}}.

\bibitem[Rai and Kim(2020)]{ref35}
Rai, A.; Kim, J.M.
\newblock A Novel Health Indicator Based on the Lyapunov Exponent, a Probabilistic Self-Organizing Map, and the Gini-Simpson Index for Calculating the RUL of Bearings.
\newblock {\em Measurement} {\bf 2020}, {\em 164},~108002.
\newblock {\url{https://doi.org/10.1016/j.measurement.2020.108002}}.

\bibitem[Yao et~al.(2021)Yao, Li, Liu, Yang, and Jia]{ref36}
Yao, D.; Li, B.; Liu, H.; Yang, J.; Jia, L.
\newblock Remaining Useful Life Prediction of Roller Bearings Based on Improved 1D-CNN and Simple Recurrent Unit.
\newblock {\em Measurement} {\bf 2021}, {\em 175},~109166.
\newblock {\url{https://doi.org/10.1016/j.measurement.2021.109166}}.

\bibitem[Wang et~al.(2021)Wang, Peng, Miao, Liu, Ayodeji, and Hao]{ref37}
Wang, H.; Peng, M.j.; Miao, Z.; Liu, Y.k.; Ayodeji, A.; Hao, C.
\newblock Remaining Useful Life Prediction Techniques for Electric Valves Based on Convolution Auto Encoder and Long Short Term Memory.
\newblock {\em ISA Transactions} {\bf 2021}, {\em 108},~333--342.
\newblock {\url{https://doi.org/10.1016/j.isatra.2020.08.031}}.

\bibitem[Nectoux et~al.(2012)Nectoux, Gouriveau, Medjaher, Ramasso, {Chebel-Morello}, Zerhouni, and Varnier]{ref38}
Nectoux, P.; Gouriveau, R.; Medjaher, K.; Ramasso, E.; {Chebel-Morello}, B.; Zerhouni, N.; Varnier, C.
\newblock PRONOSTIA : An Experimental Platform for Bearings Accelerated Degradation Tests.
\newblock In Proceedings of the IEEE International Conference on Prognostics and Health Management, PHM'12., Denver, Colorado, United States,  jun 2012; Vol. sur CD ROM, pp. 1--8.

\bibitem[Zhang et~al.(2023)Zhang, Feng, Ji, Yu, Ren, and Liu]{ref39}
Zhang, Y.; Feng, K.; Ji, J.C.; Yu, K.; Ren, Z.; Liu, Z.
\newblock Dynamic Model-Assisted Bearing Remaining Useful Life Prediction Using the Cross-Domain Transformer Network.
\newblock {\em IEEE/ASME Transactions on Mechatronics} {\bf 2023}, {\em 28},~1070--1080.
\newblock {\url{https://doi.org/10.1109/TMECH.2022.3218771}}.

\end{thebibliography}


%


\PublishersNote{}
\end{adjustwidth}
\end{document}